\documentclass[11pt]{article}
\usepackage[preprint]{acl}
\usepackage{times}
\usepackage{latexsym}
\usepackage[T1]{fontenc}
\usepackage[utf8]{inputenc}
\usepackage{microtype}
\usepackage{amsmath,amssymb}
\usepackage{booktabs}
\usepackage{multirow}
\usepackage{array}
\usepackage{graphicx}
\usepackage{url}
\usepackage{algorithm}
\usepackage{algpseudocode}
\usepackage{enumitem}
\usepackage{placeins}
\usepackage{fvextra}
\usepackage{colortbl}

\definecolor{statcolor}{HTML}{00785A}
\definecolor{figblue}{HTML}{0072B2}
\definecolor{figgreen}{HTML}{7FCDB0}
\newcommand{\Stat}[1]{\textcolor{statcolor}{#1}}
\newcommand{\Tier}[1]{\hfill\Stat{\textsf{#1}}}

\definecolor{placeholdergray}{gray}{0.55}

\newcommand{\base}{Base}
\newcommand{\cfg}[1]{\mbox{$\le$T#1}}
\newcommand{\Ind}{\mathbf{1}}
\newcommand{\sci}[2]{$#1\mathrm{e}{#2}$}
\newtheorem{lemma}{Lemma}

\title{FlashAttention with Statistics:\\ Low-Overhead Attention Features for Compact Hallucination Detectors}

\author{
 \textbf{Andrei Volodichev\textsuperscript{1}},
 \textbf{Maksim Savichev\textsuperscript{2}},
 \textbf{Nikita Sviridov\textsuperscript{3}},
 \textbf{Artem Yuliy\textsuperscript{3}},
 \textbf{Dmitry Simakov\textsuperscript{3}},
 \textbf{Alexey Zaytsev\textsuperscript{1, 3}}
 \\
 \textsuperscript{1}Applied AI Institute,
 \textsuperscript{2}Yandex,
 \textsuperscript{3}Risk AI Research Lab 
\\
  \small{    \textbf{Correspondence:} \href{mailto:a.volodichev@applied-ai.ru}{a.volodichev@applied-ai.ru}
 }
}

\begin{document}
\maketitle
\begin{abstract}
Large language models can produce fluent answers that the supplied context does not support. A compact frozen proxy model can detect such answers by reading the context and another model's response, even from a proprietary generator. Attention-based detectors use summary statistics of the proxy's attention weights, but standard pipelines materialize the attention matrix, whose memory grows quadratically with sequence length. We compute these statistics without materializing attention, by accumulation in the FlashAttention pass, sparse lookups after it, or a second tiled pass over the scores, and classify them by this path into nearly free and costlier ones. On an H100, the statistics of five published detectors add 16\% to the latency of a Qwen3-8B forward at 2K tokens, falling to 1.4\% at 35K, and store 22.5~KiB per response token. On three benchmarks, probes trained on frozen Qwen3 proxies of 0.6B to 8B parameters prove strong detectors: a 4B proxy outperforms the strongest of nine 4--9B LLM judges in F1 on two of the three, the proxies match or outperform specialized grounding checkers of up to 7B parameters, probing each proxy beats prompting it as a judge, and adding attention features significantly improves F1 in half of the 12 benchmark--proxy pairs, never significantly lowering it.
\end{abstract}

\section{Introduction}
\label{sec:intro}

\begin{figure}[t]
\centering
\includegraphics[width=\columnwidth]{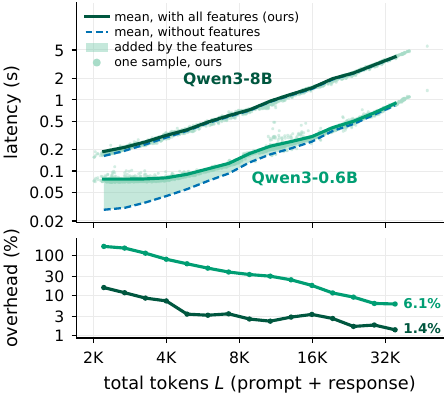}
\caption{Cost of extracting all implemented attention statistics
for hallucination detection in every layer of Qwen3-8B and
Qwen3-0.6B using our kernels.
Top: model forward-pass latency.
Bottom: relative overhead compared with the same forward using
standard scaled dot-product attention (SDPA), without statistic extraction.}
\label{fig:preview}
\end{figure}
\begin{figure*}[t]
\centering
\includegraphics[width=\textwidth]{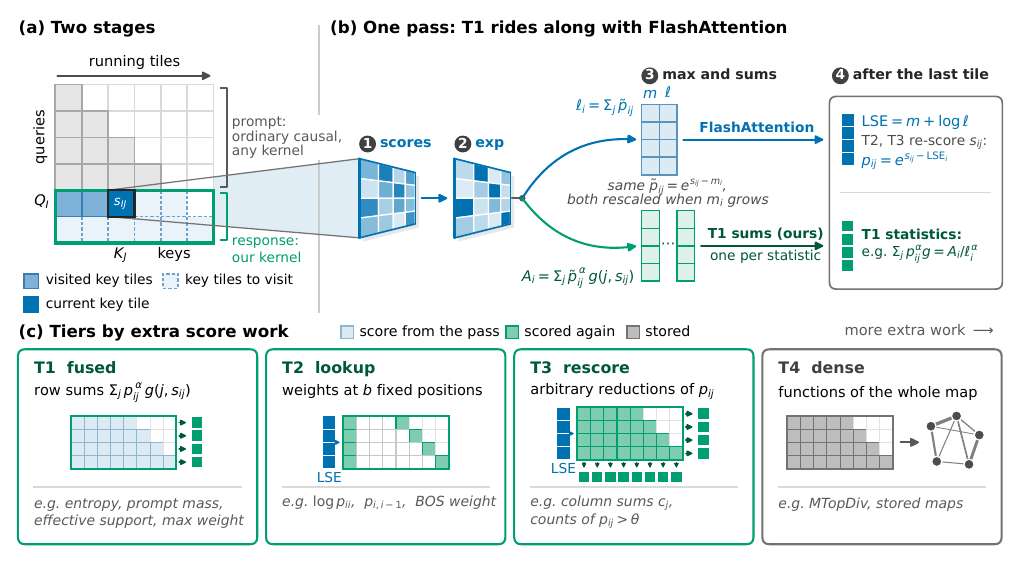}
\caption{Attention with statistics. \textcolor{figblue}{\textbf{Blue}}: standard FlashAttention \citep{dao2022flashattention,dao2024flashattention2}; \textcolor{figgreen}{\textbf{green}}: our additions. \textbf{(a)} We compute statistics for response rows only; prompt rows can go through any production kernel (\S\ref{sec:stages}). $s_{IJ}$: scores of query tile $I$ against key tile $J$. \textbf{(b)} One tile of the pass: the exponentials that FlashAttention already computes also feed the T1 statistics, which are finalized after the last tile; the pass also stores the log-sum-exp (LSE) that T2 and T3 reuse (\S\ref{sec:online-softmax}). \textbf{(c)} The four tiers, i.e.\ computation paths of a statistic, ordered by the extra work they need: none (T1, fused), a few positions per row (T2, lookup), a second pass over all scores (T3, rescore), or the whole stored map (T4, dense) (\S\ref{sec:method}).}
\label{fig:tiled-statistics}
\end{figure*}

Contextual hallucination detection asks whether an answer is supported by the supplied information. The detector need not be the model that generated the answer: the generator may expose neither its weights nor its attention, while a separate detector can be much smaller and cheaper. A compact proxy instead reads the context and completed answer and uses its own internal signals to assess context reliance. Prior work has used surrogate probabilities to estimate black-box confidence \citep{shrivastava2023llamas}, auxiliary models to calibrate generators \citep{ulmer2024apricot,kapoor2024taught,hayes2026pinocchio}, and hidden states or attention from reader models to detect hallucinations in another model's answers \citep{chwang2024androids,han2025factualityprobes,yeh2026lumina,singh2026proxyanalyzers}.

Attention-based detectors need only summary statistics of attention weights, such as prompt mass \citep{chuang2024lookback}, diagonal or previous-token weights \citep{sriramanan2024llmcheck,vazhentsev2025rauq}, sink rates \citep{binkowski2026sinkprobe}, or graph features \citep{bazarova2025toha}. Yet standard pipelines materialize the full attention matrix. For a 32-head layer, it occupies about a gigabyte at 4K tokens and tens of gigabytes at 16K--32K, although the detector keeps only a linear number of statistics.

FlashAttention computes exact attention in tiles without storing this matrix \citep{dao2022flashattention,dao2024flashattention2}. We compute detector statistics\footnote{Code: \url{https://anonymous.4open.science/r/attention-guardrails-0C16}} during or right after that pass so they inherit its linear memory and most of its speed. We classify attention statistics, including those of published detectors, by how our implementation computes them exactly without the matrix: accumulation inside the pass, a fixed number of post-pass lookups, a second pass over the scores, or dense materialization (Figure~\ref{fig:tiled-statistics}). Our Triton implementation \citep{tillet2019triton} extends the FlashAttention-2 forward to compute the statistics of five published detectors: Lookback Lens \citep{chuang2024lookback}, LLM-Check \citep{sriramanan2024llmcheck}, RAUQ \citep{vazhentsev2025rauq}, INTRYGUE \citep{kuleshova2026intrygue}, and SinkProbe \citep{binkowski2026sinkprobe}. Extracting them from every head and layer adds 16\% to a Qwen3-8B forward at 2K tokens and 1.4\% at 35K, and stores 22.5~KiB per response token (Figures~\ref{fig:preview} and~\ref{fig:preview-memory}).

Our contributions are threefold. (i) We develop a taxonomy of attention statistics based on how they can be computed exactly without materializing attention maps, identifying which come almost for free with FlashAttention and which require a second pass. This taxonomy also guides the design of new inexpensive statistics. (ii) We implement these computation paths in Triton kernels and provide an attention backend, which we release publicly, that integrates them into existing model inference pipelines.
Experiments evaluate their overhead and detection performance. (iii) We show that frozen Qwen3 proxies (0.6B--8B) can detect hallucinations in larger model's answers across three benchmarks. A 4B proxy outperforms the strongest of nine LLM judges (4B--9B) on two benchmarks, and even the smallest, 0.6B proxy beats it on the third and trails it by at most 0.03 F1 on the other two. Adding attention features significantly improves F1 in 6 of the 12 dataset--proxy pairs and never significantly lowers it, emphasizing the usefulness of attention statistics.

\section{Background}
\label{sec:background}

\subsection{Setting and notation}
\label{sec:setting}
\label{sec:stages}
We consider one head of one layer and drop both indices. Let $Q,K,V\in\mathbb{R}^{L\times d}$ be its query, key, and value projections after positional encoding; under grouped-query attention, the heads of a group share $K$ and $V$. Positions are indexed from 0. For a query row $i$, let $\mathcal J_i$ be the keys it may attend to after causal and padding masking, and define
\begin{equation}
\begin{aligned}
s_{ij}&=\tau\,q_i^\top k_j,&
\mathrm{LSE}_i&=\log\sum_{j\in\mathcal J_i}e^{s_{ij}},\\
p_{ij}&=e^{s_{ij}-\mathrm{LSE}_i},&
o_i&=\sum_{j\in\mathcal J_i}p_{ij}v_j,
\end{aligned}
\label{eq:attention}
\end{equation}
where $s_{ij}$ are the attention \emph{scores}, $p_{ij}$ the attention \emph{weights}, and $\tau=1/\sqrt d$ unless the model sets another scale. A \emph{statistic} is a per-head quantity that a kernel returns: a \emph{row reduction} of one row $p_{i\cdot}$, such as its entropy, or a \emph{column reduction} of one column $p_{\cdot j}$ over a set of rows. A kernel computes each statistic through one computation path, or \emph{tier} (Section~\ref{sec:taxonomy}). A \emph{feature} is a classifier input that the host derives from statistics, for example by averaging over rows (Appendix~\ref{app:stats}), and a \emph{detector recipe} selects features, heads, and layers for a classifier.

When the response is known in advance, as in teacher-forced detection, the input is a prompt of $P$ tokens followed by $R=L-P$ response tokens at positions $\mathcal R=\{P,\dots,L-1\}$. The detectors we target read statistics on these response rows. Under the causal mask, a prompt row $i<P$ sees only prompt keys, so its output does not depend on the response and any production kernel can compute it. The \emph{response stage} evaluates $Q_{\mathcal R}$ against all of $K$ and $V$ with a causal mask aligned to the bottom-right corner, as in key--value-cache append kernels \citep{ye2025flashinfer} (Figure~\ref{fig:tiled-statistics}a). We call its $R\times L$ block of scores and weights the \emph{response block}; without padding, it has $N_{\mathcal R}=RP+R(R+1)/2$ unmasked scores per head.

\subsection{Online softmax in FlashAttention}
\label{sec:online-softmax}
FlashAttention computes $o_i$ in tiles without storing the weights \citep{dao2022flashattention,dao2024flashattention2}. The FlashAttention-2 forward pass is query-major: a tile $I$ of query rows stays resident while the key tiles $J$ it sees stream past (Figure~\ref{fig:tiled-statistics}b). Its backward pass is key-major: a key tile stays resident while query tiles stream past. We call one evaluation of every score of the response block, in tiles, a \emph{pass}; \emph{the pass} is the query-major response stage.

For one row, online softmax carries a running maximum $m$, a normalizer $\ell$, and an unnormalized output $O$ \citep{milakov2018online,rabe2021selfattention}. For a new key tile $J$, let $m'=\max(m,\max_{j\in J}s_j)$, $a=e^{m-m'}$, and $\tilde p_j=e^{s_j-m'}$. The update is
\begin{equation}
\begin{aligned}
\ell&\leftarrow a\ell+\sum_{j\in J}\tilde p_j,\\
O&\leftarrow aO+\sum_{j\in J}\tilde p_jv_j,\qquad m\leftarrow m'.
\end{aligned}
\label{eq:online-softmax}
\end{equation}
After the last tile, $o=O/\ell$ and $\mathrm{LSE}=m+\log\ell$. As FlashAttention stores the LSE, a later kernel can recover any weight (Equation~\ref{eq:attention}) by re-evaluating its score.

\begin{table}[t]
\centering
\small
\setlength{\tabcolsep}{3pt}
\newcommand{\TierEx}[1]{\newline{\scriptsize\itshape #1\par}}
\begin{tabular}{@{}>{\raggedright\arraybackslash}p{3.4cm}lll@{}}
\toprule
Tier & Scores & Stat.\ ops & Memory\\
\midrule
T1 (fused)\TierEx{entropy, prompt mass, effective support, max weight} & none & $O(N_sN_{\mathcal R})$ & $N_sR$\\
\addlinespace[2pt]
T2 (lookup)\TierEx{$\log p_{ii}$, $p_{i,i-1}$, BOS weight} & $bR$ & $O(bR)$ & $bR$\\
\addlinespace[2pt]
T3 (rescore)\TierEx{column sums, sink rates, threshold counts} & $N_{\mathcal R}$ & $O(N_{\mathcal R})$ & $O(R+L)$\\
\addlinespace[2pt]
T4 (dense)\TierEx{MTopDiv, stored maps} & $N_{\mathcal R}$ & $\Omega(N_{\mathcal R})$ & $N_{\mathcal R}$\\
\bottomrule
\end{tabular}
\caption{Work per head beyond the pass over the response block, by tier. \textbf{Scores}: extra attention scores evaluated, each costing $d$ multiply-adds and one exponential. \textbf{Stat.\ ops}: statistic-specific elementwise operations, plus the statistic's own algorithm for T4. \textbf{Memory}: numbers written per head besides the attention output and LSE.}
\label{tab:tier-paths}
\end{table}

\section{Computing Attention Statistics without Materialization}
\label{sec:method}

\subsection{Computation paths}
\label{sec:taxonomy}
We compute statistics only in the response stage (Section~\ref{sec:stages}, Figure~\ref{fig:tiled-statistics}a). A statistic either fits a bounded state that the pass can update, or it needs final weights, which a later kernel recovers from a re-evaluated score and the stored LSE (Section~\ref{sec:online-softmax}). We sort statistics into four tiers by the scores evaluated beyond the pass (Table~\ref{tab:tier-paths}, Figure~\ref{fig:tiled-statistics}c): \textbf{T1 (fused)} evaluates none, \textbf{T2 (lookup)} a fixed number $b$ per row, \textbf{T3 (rescore)} every score a second time, and \textbf{T4 (dense)} materializes the response block. A tier is the path our implementation uses, with Triton kernels for T1--T3 and eager PyTorch for T4, so it bounds the cost from above: the pass could also compute the T2 lookups (Section~\ref{sec:post-pass}), and T1 holds only the bounded-state reductions we implement. The LSE, which every FlashAttention forward writes, belongs to no tier. Appendix~\ref{app:stats} defines every statistic we compute. Appendix~\ref{app:impl} gives the pseudocode and kernel details.

\subsection{T1 (fused): statistics in online softmax}
\label{sec:accumulator}
\textbf{T1 (fused)} statistics are accumulated inside the pass and need no extra scores. Consider prompt mass $\sum_{j<P}p_{ij}$, the share of attention that row $i$ puts on the prompt, from which Lookback Lens builds its features \citep{chuang2024lookback}. During the pass, row $i$ knows each weight only up to the factor $e^{m_i-\mathrm{LSE}_i}$, which is fixed only after its last key tile. Prompt mass can nevertheless be accumulated like $\ell$: keep $A=\sum_{j<P}e^{s_j-m}$ over the keys seen so far, update it as $A\leftarrow aA+\sum_{j\in J,\,j<P}\tilde p_j$ next to Equation~\ref{eq:online-softmax}, and output $A/\ell$ after the last tile.

The same works for integer powers of the weights times any function of the score. For an integer $\alpha_u\ge1$ and a fixed-dimensional function $g_u$ that does not depend on $m$, $\ell$, or the LSE, keep the state $A_u$ and update it for each key tile:
\begin{align}
A_u&=\sum_{j\ \mathrm{seen}}e^{\alpha_u(s_j-m)}\,g_u(i,j,s_j),
\label{eq:weighted-state}\\
A_u&\leftarrow a^{\alpha_u}A_u+\sum_{j\in J}\tilde p_j^{\,\alpha_u}g_u(i,j,s_j).
\label{eq:update}
\end{align}
Since $\alpha_u$ is an integer, $\tilde p_j^{\,\alpha_u}$ is a product of exponentials the pass already computes. This standard consequence of online softmax defines T1:

\begin{lemma}[Separable row reductions]
\label{lem:separable}
For $N_s$ statistics, let $g_1,\dots,g_{N_s}$ be functions as above and $\alpha_1,\dots,\alpha_{N_s}\ge1$ integers. The sums
\begin{equation}
F_i^{(u)}=\sum_{j\in\mathcal J_i}p_{ij}^{\alpha_u}\,g_u(i,j,s_{ij}),
\label{eq:general-stat}
\end{equation}
and any fixed function of $F_i^{(1)},\dots,F_i^{(N_s)}$ and $\mathrm{LSE}_i$, can be computed in the pass, with no additional exponentials or scores and $\sum_u\dim g_u$ numbers of state per row, independent of $L$.
\end{lemma}
Prompt mass is the case $\alpha=1$, $g=\Ind[j<P]$; entropy is $H=\mathrm{LSE}-F$ with $\alpha=1$ and $g=s$; effective support is $1/F$ with $\alpha=2$ and $g=1$; and the maximum weight is $1/\ell$. The kernel updates these accumulators per score tile next to the FlashAttention state, keeping nothing per key (Figure~\ref{fig:tiled-statistics}b). Appendix~\ref{app:separable} gives the proof and the accumulator's merge and stability properties.

\begin{figure*}[t]
\centering
\includegraphics[width=\textwidth]{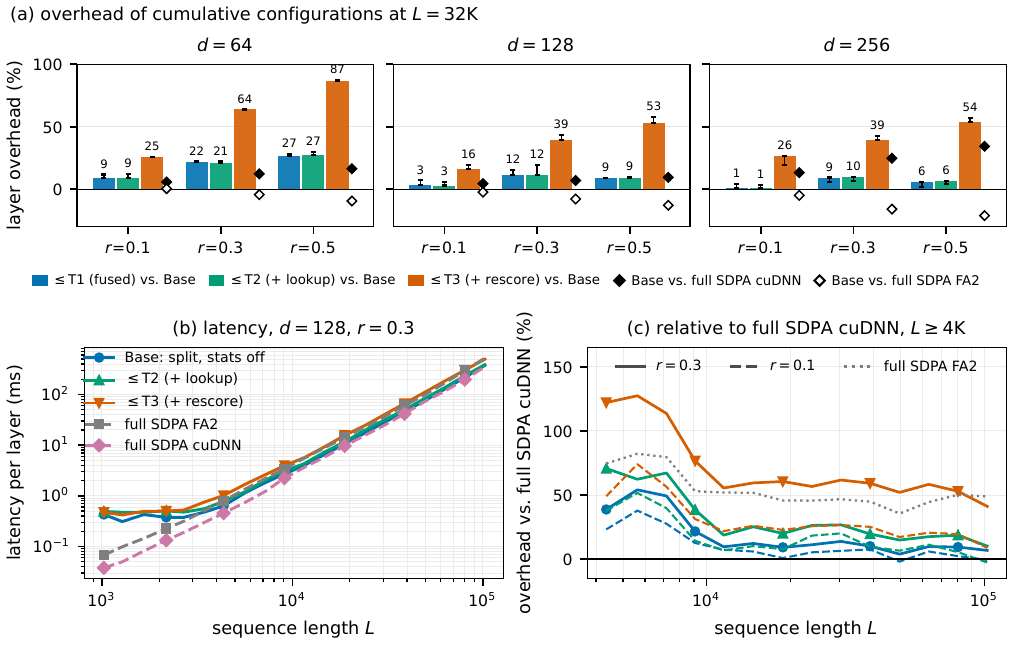}
\caption{What each benchmark configuration adds to one attention layer (Q2). Configuration \cfg{k} enables the paths up to T$k$; \cfg{3} reads the BOS column sum from the column sums instead of the lookup. \textbf{(a)} Overhead of each configuration over our statistics-off \base{} layer at $L{=}32\mathrm{K}$ (whiskers: 16K--32K), and of \base{} relative to SDPA on the whole sequence (diamonds). \textbf{(b)} Latency from 1K to 100K tokens at $d=128$, $r=0.3$. \textbf{(c)} Overhead relative to SDPA cuDNN on the whole sequence, at $r=0.3$ (solid) and $r=0.1$ (dashed).}
\label{fig:tier-cost}
\end{figure*}

\subsection{T2--T4: post-pass statistics}
\label{sec:post-pass}
\paragraph{T2 (lookup).}
Some statistics read a fixed number $b$ of weights per response row at positions known in advance, optionally summed over rows: the log-diagonal $\log p_{ii}=s_{ii}-\mathrm{LSE}_i$, the previous-token weight $p_{i,i-1}$, and the weight on a fixed sink column such as the beginning-of-sequence (BOS) token \citep{sriramanan2024llmcheck,vazhentsev2025rauq,xiao2024streamingllm}. The pass could keep these $b$ scores per row and finalize them with the LSE, which would place them in T1. We instead compute them in a separate kernel that reads only $Q$, $K$, and the LSE, so it works after any forward kernel that stores the LSE and takes its positions at run time (Appendix~\ref{app:impl}). Its $bR$ extra scores grow linearly with the sequence length.

\paragraph{T3 (rescore).}
Column reductions, such as column entropy (not implemented) and the column sums $c_j=\sum_{i\in\mathcal R}p_{ij}$ from which sink rates are read \citep{gu2025attentionsink,binkowski2026sinkprobe}, cannot be finalized by the query-major pass from bounded state per query row. Row $i$ contributes $e^{s_{ij}-m_i}$ to column $j$, and the factor that turns this into $p_{ij}$ is known only after the row's last key tile and differs between rows (Section~\ref{sec:accumulator}), so a column sum needs each score again once its row's LSE is final, or all scores stored. The same holds for row reductions that compare each weight with a function of $\mathrm{LSE}_i$, such as counts of $p_{ij}>\theta$ (not implemented). A second kernel re-evaluates each score, converts it with the stored LSE, and reuses the key-major loop of the FlashAttention-2 backward pass (Appendix~\ref{app:impl}). It evaluates all the exponentials of the response stage, which run on slow non-tensor-core units \citep{shah2024flashattention3}, but only half its multiply-adds (no product with $V$), so it costs close to a second response stage (Figure~\ref{fig:tier-cost}a).

\paragraph{T4 (dense).}
Statistics that are not bounded-state reductions of the weights, such as the attention-graph divergence MTopDiv \citep{bazarova2025toha} and stored attention maps, need the materialized response block, with $O(N_{\mathcal R})$ memory; we have no T4 kernel.

\section{Kernel Benchmark}
\label{sec:kernel-eval}
We evaluate the kernels on their own and ask three questions. (Q1) Are the fused statistics numerically at least as accurate as those from materialized attention? (Q2) What does each tier add on top of attention itself, and how does that compare with production kernels? (Q3) How much does a whole-model forward slow down when it computes the statistics underlying several published detectors with our kernels, and how does that compare with computing them from materialized attention?

\paragraph{Setup.}
All runs use a single NVIDIA H100 NVL in bf16. Each of our configurations is a whole attention layer: prompt rows go to PyTorch's SDPA and response rows to our Triton kernels. \base{} is the same layer with statistics off. We compare both with SDPA on the whole sequence (``full SDPA'') with the FlashAttention-2 and cuDNN backends. See Appendix~\ref{app:setup} for software versions and the timing protocol, and Appendix~\ref{app:memory} for memory.

\paragraph{Benchmark configurations.}
We measure the tiers in cumulative configurations: \cfg{k} enables the paths up to T$k$ on top of \base{}, which writes only the output and the LSE. \cfg{1} adds the T1 row statistics of Table~\ref{tab:tier-paths}, \cfg{2} the lookups of $\log p_{ii}$, $p_{i,i-1}$, and the BOS weight $p_{is}$, and \cfg{3} the column sums, which replace the BOS lookup by the BOS column sum $c_s$, so \cfg{3} does not return the per-row $p_{is}$. All configurations run on the $R=rL$ response rows, so their cost grows with $r$.

\paragraph{Numerical accuracy (Q1).}
We measure the absolute error on response rows against an fp64 reference, for materialized attention and for our kernels, on random inputs and on post-RoPE $Q,K,V$ captured from Qwen3-8B, from 2K to 32K tokens. \textbf{Finding.} Our statistics are three to four orders of magnitude more accurate for every statistic (Table~\ref{tab:numerics} in Appendix~\ref{app:numerics}), because our kernels reduce fp32 probabilities in registers, whereas materialized attention first rounds each to bf16.

\paragraph{Marginal cost per tier (Q2).}
We time one attention layer at batch 1 on synthetic inputs of $L\in\{4\mathrm{K},16\mathrm{K},32\mathrm{K}\}$ tokens with response fraction $r=R/L\in\{0.1,0.3,0.5\}$, in Qwen3-8B shapes (32 query heads, 8 KV heads, $d=128$) and at $d=64$ and $256$. \cfg{2} adds at most 5 percentage points of compute time over \cfg{1}, and both add 1--30\% over \base{} (Figure~\ref{fig:tier-cost}a; Table~\ref{tab:tiers-cost} in Appendix~\ref{app:cost}). The overhead generally increases with $r$ and decreases with $d$, both trends explained by T1: the former reflects the $R=rL$ response rows processed, while the latter reflects matrix computation scaling with $d$ versus T1's elementwise work. \cfg{3} is the costliest configuration, at 16--87\%, and shows the same scaling in $r$ as \cfg{1}; in $d$, it drops from $d=64$ to $d=128$ but not further, because T3's second pass adds its own matrix work.

Our kernels scale like the production ones all the way to 100K tokens (Figure~\ref{fig:tier-cost}b--c), with no wall expected further: at 100K tokens and $r=0.3$, a layer with all row statistics takes 387~ms, against 351~ms for cuDNN and 523~ms for FA2 without statistics. Below about 4K tokens, our latency stops falling because a fixed CPU overhead per call dominates (Appendix~\ref{app:cost}). In memory, the statistics take 7.6--20~MiB of a layer at 32K (Appendix~\ref{app:memory}). \textbf{Finding.} T1 and T2 are cheap enough to compute together by default; T3 is worthwhile only when a detector needs reductions over all final weights.

\begin{figure}[t]
\centering
\includegraphics[width=\columnwidth]{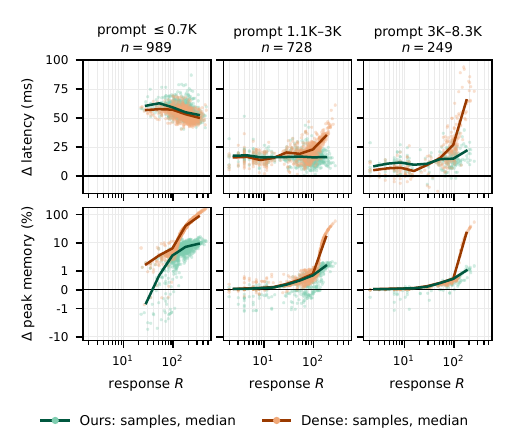}
\caption{Added latency and peak memory over the forward without features, per sample, by prompt length (Q3). Lines: binned medians. The shortest prompts are RAGTruth QA, the others PHANTOM.}
\label{fig:q3-scaling}
\end{figure}

\paragraph{Comparison with prior methods (Q3).}
We compute the attention statistics underlying Lookback Lens (T1) \citep{chuang2024lookback}, LLM-Check (T2) \citep{sriramanan2024llmcheck}, RAUQ (T2) \citep{vazhentsev2025rauq}, INTRYGUE (T3) \citep{kuleshova2026intrygue}, and SinkProbe (T3) \citep{binkowski2026sinkprobe}, with the other T1 statistics, for all 32 heads of all 36 layers of Qwen3-8B, in one teacher-forced forward at batch 1. Where a method pools over all tokens, as LLM-Check and SinkProbe do, our response-row version is an adaptation (Appendix~\ref{app:stats}). The floor is the same forward without statistics. We stop at feature extraction; the classifiers cost little and the same for every path. See Appendix~\ref{app:preview} for the SDPA backends and the dense baseline.

\textbf{Data.} From RAGTruth \citep{niu2024ragtruth}, we take the subset of QA responses generated by Llama-2-70B-chat, and from PHANTOM \citep{phantom2025}, the 2,000-token split of the 10-K filings set (Appendix~\ref{app:preview}). Inputs are short here because dense pipelines do not scale much further.

\textbf{Results.} Figure~\ref{fig:q3-scaling} shows the cost of every sample, grouped by prompt length. At these lengths, both paths add about the same latency for every prompt length, but dense latency starts to grow once a response exceeds about 100 tokens, and faster for longer prompts: on the sample with the most response weights, 785 response tokens in 7.2K, dense adds 501~ms to the forward and ours 53~ms. \textbf{Finding.} The clear difference is memory: dense memory grows with $RL$, by 2.1~GiB on that sample, whereas ours grows only with $R$, by 23~MiB, so the gap widens with sequence length.

\section{Proxy Detection}
\label{sec:evaluation}
\label{sec:detection-eval}
Section~\ref{sec:kernel-eval} showed that attention statistics are cheap to extract; here we ask whether a compact proxy detector should extract them. Every proxy forward already yields hidden states and logits at no extra cost, and probes on hidden states are strong detectors on their own \citep{azaria2023internal,chen2024inside}, so attention statistics earn their place only if they add to these signals. We ask four questions. (Q4) Do attention features improve a detector built on hidden states and logits? (Q5) Which attention representation helps: a few light statistics pooled over selected heads, from T1 and T2, or heavier sink statistics of every head, which need T3? (Q6) How does detection scale with proxy size? (Q7) How do proxy detectors compare with LLM judges? Section~\ref{sec:extraction-cost} then gives the cost of the features in the detection pipeline, and Appendix~\ref{app:transfer} tests whether published attention detectors, computed from one forward of the 8B proxy, transfer across the six domains of HaluBench.

\begin{table*}[t]
\centering
\small
\setlength{\tabcolsep}{3.6pt}
\begin{tabular}{lr*{12}{c}}
\toprule
& & \multicolumn{4}{c}{RAGTruth QA} & \multicolumn{4}{c}{HaluBench} & \multicolumn{4}{c}{TRIVIA+}\\
\cmidrule(lr){3-6}\cmidrule(lr){7-10}\cmidrule(lr){11-14}
Features & $N$ & 0.6B & 1.7B & 4B & 8B & 0.6B & 1.7B & 4B & 8B & 0.6B & 1.7B & 4B & 8B\\
\midrule
A$_7$ & 7 & .573 & .567 & .503 & .550 & .615 & .639 & .631 & .637 & .531 & .524 & .545 & .527\\
\addlinespace[1pt]
L$_9$ & 9 & .440 & .444 & .467 & .433 & .633 & .627 & .642 & .665 & .545 & .523 & .518 & .576\\
A$_7$L$_9$ & 16 & .566 & .532 & .541 & .546 & .642 & .643 & .656 & .675 & .555 & .560 & .543 & .578\\
\addlinespace[1pt]
H$_{64}$ & 64 & .557 & .559 & \underline{.566} & .557 & .697 & .734 & .742 & .742 & .627 & .638 & .634 & \underline{.670}\\
A$_7$H$_{64}$ & 71 & \textbf{.637} & \textbf{.581} & .547 & \textbf{.600} & .698 & .729 & .744 & .746 & .613 & .638 & .639 & .645\\
\addlinespace[1pt]
H$_{64}$L$_9$ & 73 & .614 & .559 & .543 & .533 & .697 & \underline{.736} & .749 & .754 & .599 & \underline{.641} & .641 & .647\\
A$_7$H$_{64}$L$_9$ & 80 & \underline{.621} & .563 & .528 & .563 & .702 & .726 & .755 & .756 & \textbf{.641} & \textbf{.643} & \underline{.646} & .667\\
\midrule
A$_{64}$ & 64 & .593 & \underline{.577} & \textbf{.583} & .592 & \textbf{.727} & .718 & \underline{.769} & \underline{.768} & .631 & .634 & .616 & .656\\
A$_{64}$H$_{64}$ & 128 & .601 & .529 & \underline{.566} & \underline{.594} & \underline{.724} & \textbf{.746} & \textbf{.776} & \textbf{.791} & \underline{.636} & .640 & \textbf{.674} & \textbf{.671}\\
\midrule
\multicolumn{2}{@{}l}{Best LLM judge} & \multicolumn{4}{c}{.623} & \multicolumn{4}{c}{.755} & \multicolumn{4}{c}{.655}\\
\bottomrule
\end{tabular}
\caption{Test F1 by feature set and Qwen3 proxy size. A$_7$: cheap attention features; A$_{64}$: all-head sink values; H$_{64}$: compact hidden states; L$_9$: logit features; combinations concatenate blocks (Section~\ref{sec:in-domain}). $N$: number of classifier inputs. \textbf{Bold}: best in column; \underline{underlined}: second best. Bottom: the best of nine LLM judges of 4--9B parameters, Qwen3.5-9B on every dataset (Table~\ref{tab:judges}).}
\label{tab:det-f1}
\end{table*}

\subsection{Task, data, and metrics}
\label{sec:eval-setup}
\label{sec:proxy}
A detector reads a context, a question, and an existing answer and predicts whether the context supports the answer. A frozen Qwen3 proxy \citep{yang2025qwen3} of 0.6B, 1.7B, 4B, or 8B parameters reads each example in one teacher-forced forward, so it need not be the model that wrote the answer.

We use RAGTruth QA \citep{niu2024ragtruth}, HaluBench \citep{ravi2024lynx}, and TRIVIA+ \citep{chen2026triviaplus}, split into training, validation, and test so that groups of related examples never cross splits. We filter and split HaluBench ourselves, because on its unfiltered answers a TF--IDF classifier given only the answer reaches 0.84 AUROC. Appendix~\ref{app:data} gives the splits, the filtering, and text-only baselines.

Table~\ref{tab:det-f1} reports test F1 with validation-selected threshold, because a deployed detector must label each answer; AUROC is in Table~\ref{tab:det-auc} of Appendix~\ref{app:detection}.

\subsection{Feature blocks}
\label{sec:in-domain}
One proxy forward yields three feature blocks: hidden states and logits from the forward pass, and attention features added by our kernels in two variants, A$_7$ and A$_{64}$. Feature sets concatenate one or more blocks (e.g., A$_7$H$_{64}$). We select the A$_7$ and L$_9$ features and the H$_{64}$ pooling and projection recipe once, using the highest mean AUROC on the TRIVIA+ and reuse it across all experiments with the details of the protocol in Appendix~\ref{app:detection}. Each per-token quantity, i.e., an attention-head statistic, a logit statistic, or a hidden-state coordinate, is pooled over answer tokens by its mean, standard deviation, minimum, and maximum, producing distinct ppoled features.

\textbf{Cheap attention features (A$_7$)} use the $K$ strongest induction heads, which attend to earlier input tokens \citep{olsson2022induction}. 
Each head yields 22 candidate features, averaged across the $K$ heads: four pooled features for each of five per-token statistics, the T2 BOS sink rate, and the log-diagonal. Feature selection retains the seven in bold: maximum weight (\textbf{mean}, \textbf{std}, min, \textbf{max}), effective support (mean, std, \textbf{min}, max), entropy (\textbf{mean}, \textbf{std}, min, max), the lookback ratio built from prompt mass and the previous-token weight (mean, std, min, max), the \textbf{BOS sink rate}, and the log-diagonal. 

\textbf{All-head sink values (A$_{64}$)} are the richest attention features our kernels provide. Following SinkProbe \citep{binkowski2026sinkprobe}, we keep, for every head of every layer, its $k$ largest sink values from T3, the normalized column sums $\bar c_j$ (Appendix~\ref{app:stats}), concatenate them, and project them onto 64 principal components, matching the size of H$_{64}$ without selection. 

\textbf{Compact hidden states (H$_{64}$)} are the standard probe input. Following prior work \citep{azaria2023internal,chen2024inside}, we use one middle layer without selection. The selection keeps the stacked mean and standard deviation of its coordinates, which are standardized and projected onto 64 principal components (Appendix~\ref{app:recipes}).

\textbf{Logit features (L$_9$)} are nine of 224 candidates from seven per-token statistics of the next-token distribution, e.g., the answer token's minimum probability, kept by recursive feature elimination (Appendix~\ref{app:logits}).

\subsection{Selection and training per dataset and proxy}
\label{sec:selection}
Each dataset--proxy pair refits the scalers and projections on its training split and selects hyperparameters without test labels. For A$_7$, the number of heads $K\in\{1,2,4,8,16,32\}$ is chosen by the validation AUROC of A$_7$ alone and reused in every set that contains A$_7$. For A$_{64}$, the number of sink values per head $k\in\{1,\dots,10\}$ is chosen for each pair by grouped five-fold cross-validation on its training split and reused in A$_{64}$H$_{64}$. Every feature set serves as an input to a class-balanced $L_2$ regularized logistic regression with regularization strength chosen by validation AUROC. 

\subsection{Main results}
\label{sec:sink64}
\label{sec:scaling}
\label{sec:judges}
\textbf{Attention features (Q4).} In every one of the 12 dataset--proxy pairs, the feature set with the best F1 contains A$_7$ or A$_{64}$ (Table~\ref{tab:det-f1}), and so does the set with the best AUROC (Table~\ref{tab:det-auc}). 
With only 150--546 test groups, uncertainty in F1
differences remains substantial. After Holm correction,
adding $A_7$ significantly improves F1 for all four
proxies on RAGTruth QA when combined with logits,
and for two proxies when combined with hidden states (Table~\ref{tab:det-sig}).
Adding $A_{64}$ to hidden states yields significant
improvements for the 4B and 8B proxies on HaluBench (Table~\ref{tab:sink-sig}).
Thus, overall attention features provide improvements if compared to variants without them.

\textbf{Which attention features (Q5).} Alone, A$_{64}$ beats A$_7$ in all 12 pairs in both F1 and AUROC, which supports the heavier T3 path when attention is the main signal, although A$_{64}$ also has more inputs and covers every head. With hidden states, A$_{64}$H$_{64}$ wins on HaluBench and TRIVIA+ with every proxy, while on RAGTruth QA A$_7$H$_{64}$ wins with three of four. The cheap features thus remain useful on some datasets  alongside T3 sinks.

\textbf{Proxy size (Q6).} 
The benefit of larger proxies depends on the benchmark and feature representation. On HaluBench
and TRIVIA+, larger models improve detectors using hidden states, logits, or all-head sink features. 
On RAGTruth QA, even the smallest proxy remains competitive. 
These results favour selecting proxy size and features jointly on validation data with small 0.6B proxy showing strong performance in multiple settings.

\textbf{LLM judges (Q7).}
Table~\ref{tab:judges} compares our proxy detectors with nine
instruction-tuned LLM judges of 4--9B parameters
\citep{zheng2023judging}; evaluation details are in
Appendix~\ref{app:judges}. Against Qwen3.5-9B, the strongest
evaluated judge, a 4B proxy achieves higher observed F1 on
HaluBench and TRIVIA+, and a 0.6B proxy does so on RAGTruth QA.
Our probes are also competitive with the specialized grounding
checkers MiniCheck-7B \citep{tang2024minicheck} and HHEM-2.1
\citep{li2024hhem}. 
The most consistent advantage is within the same backbone:
A$_{64}$H$_{64}$ achieves higher F1 and AUROC than prompting
the corresponding Qwen3 model in all 12 dataset--proxy pairs.
When task-specific training data are
available, supervised probing therefore offers an effective
way to build compact detectors that outperform prompting
the same models and compete with larger judges.

\subsection{Cost of feature extraction}
\label{sec:extraction-cost}
A proxy detector adds a linear classifier to one proxy forward, so it costs one forward with features. Figure~\ref{fig:preview} times forwards of the largest and smallest proxies, Qwen3-8B and Qwen3-0.6B, with all kernel statistics for every head and layer, against the same forwards without features (Appendix~\ref{app:preview}). From 2K to 35K tokens, the overhead falls from 16\% to 1.4\% for Qwen3-8B and from about 170\% to 6\% for Qwen3-0.6B, whose whole forward at 2K still takes under 100~ms. On shorter inputs, Python-side CPU overhead sets the floor. In peak memory \emph{above the model weights} (Figure~\ref{fig:preview-memory}), the features add about 28~KiB per response token for Qwen3-8B and 11~KiB for Qwen3-0.6B, mostly the stored statistics (22.5 and 8.75~KiB; Appendix~\ref{app:memory}).

\section{Conclusion}
\label{paper:main-end}

Attention statistics are an active basis for hallucination detection \citep{chuang2024lookback,binkowski2026sinkprobe,kuleshova2026intrygue}.
We confirmed that many those statistics can
be extracted without materializing attention matrices. 
Our paper classifies these statistics by how they are computed, clarifying
which can reuse the attention pass and which require additional
lookups or a second pass over the scores. Our
FlashAttention-compatible backend implements these paths with
memory linear in sequence length. In our H100 NVL benchmarks
at batch size one, extracting the statistics underlying five
detectors adds just 16\% to Qwen3-8B forward latency at 2K
tokens and 1.4\% at 35K. This implementation supports attention
features used by existing detectors and their use with small,
frozen proxy models. Our experiments show that supervised
detectors built on these proxies can compete with larger
prompted judges. 
Reducing Python overhead and synchronization with the GPU would further improve extraction speed on short inputs.

\clearpage
\section*{Limitations}
We do not claim that our tiers list every attention statistic that can be computed efficiently. A tier is the path our implementation uses (Section~\ref{sec:taxonomy}): a statistic in T4 may admit an efficient exact kernel that we did not write, and we do not claim that our algorithms are optimal.

We write our kernels in Triton, which is portable across GPUs but can be slower than specialized CUDA kernels: our response-stage kernel is faster than FA2 on Hopper but slower than cuDNN. Nothing prevents computing the same statistics inside such a kernel. Our Python wrapper also adds a fixed host cost of 47--60~ms per forward. Little of it is GPU work, which grows by only 5~ms on short Qwen3-8B inputs (Appendix~\ref{app:preview}), so a compiled host implementation with CUDA graphs could remove most of it.

Our implementation is not yet a production kernel. It computes the statistics of a complete response in one call. Computing them token by token during generation would need a split-KV decoding kernel, which is possible because the T1 accumulators merge associatively, but which we have not implemented. It supports neither sliding-window nor soft-capped attention, takes only a padding mask, and covers head dimensions 64, 128, and 256, so it does not yet serve an arbitrary model. Our wrapper also runs the prompt stage in the same layer call and copies both stage outputs into one tensor. This copy does not raise the peak of a whole forward (Appendix~\ref{app:memory}) and is avoidable: since our statistics cover response rows only, the prompt can go through a separate prefill whose key--value cache the response stage reads, as serving systems do \citep{kwon2023pagedattention}.

The proxy requires an auxiliary forward pass and labeled probe-training examples. A small proxy may miss evidence available to a larger generator. We assess support relative to supplied context, not general factual truth. Scores require separate calibration before interpretation as error probabilities. Dataset artifacts, label disagreements, truncation, and domain shift can affect results.

\section*{Ethical Considerations}
Our work aims to make hallucination detection cheap enough to check every response of a retrieval-augmented system. Like any detector, it can miss unsupported answers or flag supported ones, so in high-stakes settings it should complement human review rather than replace it. All experiments use public benchmarks under their terms of use; we collect no new data and involve no human subjects.

\bibliography{references}

@inproceedings{dao2022flashattention,
  author = {Dao, Tri and Fu, Daniel Y. and Ermon, Stefano and Rudra, Atri and R{\'e}, Christopher},
  title = {{FlashAttention: Fast and Memory-Efficient Exact Attention with IO-Awareness}},
  booktitle = {Advances in Neural Information Processing Systems},
  volume = {35},
  pages = {16344--16359},
  year = {2022},
  url = {https://papers.nips.cc/paper_files/paper/2022/hash/67d57c32e20fd0a7a302cb81d36e40d5-Abstract-Conference.html},
  doi = {10.52202/068431-1189}
}

@inproceedings{dao2024flashattention2,
  author = {Dao, Tri},
  title = {{FlashAttention-2: Faster Attention with Better Parallelism and Work Partitioning}},
  booktitle = {The Twelfth International Conference on Learning Representations},
  pages = {35549--35562},
  year = {2024},
  url = {https://proceedings.iclr.cc/paper_files/paper/2024/hash/98ed250b203d1ac6b24bbcf263e3d4a7-Abstract-Conference.html}
}

@inproceedings{shah2024flashattention3,
  author = {Shah, Jay and Bikshandi, Ganesh and Zhang, Ying and Thakkar, Vijay and Ramani, Pradeep and Dao, Tri},
  title = {{FlashAttention-3: Fast and Accurate Attention with Asynchrony and Low-precision}},
  booktitle = {Advances in Neural Information Processing Systems},
  volume = {37},
  pages = {68658--68685},
  year = {2024},
  url = {https://proceedings.neurips.cc/paper_files/paper/2024/hash/7ede97c3e082c6df10a8d6103a2eebd2-Abstract-Conference.html},
  doi = {10.52202/079017-2193}
}

@article{rabe2021selfattention,
  author = {Rabe, Markus N. and Staats, Charles},
  title = {{Self-attention Does Not Need $O(n^2)$ Memory}},
  journal = {arXiv preprint arXiv:2112.05682},
  year = {2021},
  url = {https://arxiv.org/abs/2112.05682},
  doi = {10.48550/arXiv.2112.05682}
}

@article{feng2024attentionrnn,
  author = {Feng, Leo and Tung, Frederick and Hajimirsadeghi, Hossein and Ahmed, Mohamed Osama and Bengio, Yoshua and Mori, Greg},
  title = {{Attention as an {RNN}}},
  journal = {arXiv preprint arXiv:2405.13956},
  year = {2024},
  url = {https://arxiv.org/abs/2405.13956},
  doi = {10.48550/arXiv.2405.13956}
}

@inproceedings{liu2024ringattention,
  author = {Liu, Hao and Zaharia, Matei and Abbeel, Pieter},
  title = {{RingAttention with Blockwise Transformers for Near-Infinite Context}},
  booktitle = {The Twelfth International Conference on Learning Representations},
  year = {2024},
  pages = {3992--4008},
  url = {https://proceedings.iclr.cc/paper_files/paper/2024/hash/1119587863e78451f080da2a768c4935-Abstract-Conference.html}
}

@inproceedings{kwon2023pagedattention,
  author = {Kwon, Woosuk and Li, Zhuohan and Zhuang, Siyuan and Sheng, Ying and Zheng, Lianmin and Yu, Cody Hao and Gonzalez, Joseph E. and Zhang, Hao and Stoica, Ion},
  title = {{Efficient Memory Management for Large Language Model Serving with {PagedAttention}}},
  booktitle = {Proceedings of the 29th Symposium on Operating Systems Principles},
  year = {2023},
  pages = {611--626},
  url = {https://dl.acm.org/doi/10.1145/3600006.3613165},
  doi = {10.1145/3600006.3613165}
}

@inproceedings{agrawal2024sarathi,
  author = {Agrawal, Amey and Kedia, Nitin and Panwar, Ashish and Mohan, Jayashree and Kwatra, Nipun and Gulavani, Bhargav S. and Tumanov, Alexey and Ramjee, Ramachandran},
  title = {{Taming Throughput-Latency Tradeoff in {LLM} Inference with {Sarathi-Serve}}},
  booktitle = {18th USENIX Symposium on Operating Systems Design and Implementation (OSDI 24)},
  pages = {117--134},
  year = {2024},
  url = {https://www.usenix.org/conference/osdi24/presentation/agrawal}
}

@inproceedings{ye2025flashinfer,
  author = {Ye, Zihao and Chen, Lequn and Lai, Ruihang and Lin, Wuwei and Zhang, Yineng and Wang, Stephanie and Chen, Tianqi and Kasikci, Baris and Grover, Vinod and Krishnamurthy, Arvind and Ceze, Luis},
  title = {{{FlashInfer}: Efficient and Customizable Attention Engine for {LLM} Inference Serving}},
  booktitle = {Proceedings of Machine Learning and Systems},
  volume = {7},
  year = {2025},
  url = {https://proceedings.mlsys.org/paper_files/paper/2025/hash/dbf02b21d77409a2db30e56866a8ab3a-Abstract-Conference.html}
}

@inproceedings{zhang2023h2o,
  author = {Zhang, Zhenyu and Sheng, Ying and Zhou, Tianyi and Chen, Tianlong and Zheng, Lianmin and Cai, Ruisi and Song, Zhao and Tian, Yuandong and R{\'e}, Christopher and Barrett, Clark and Wang, Zhangyang and Chen, Beidi},
  title = {{{H$_2$O}: Heavy-Hitter Oracle for Efficient Generative Inference of Large Language Models}},
  booktitle = {Advances in Neural Information Processing Systems},
  volume = {36},
  pages = {34661--34710},
  year = {2023},
  url = {https://proceedings.neurips.cc/paper_files/paper/2023/hash/6ceefa7b15572587b78ecfcebb2827f8-Abstract-Conference.html},
  doi = {10.52202/075280-1506}
}

@inproceedings{li2024snapkv,
  author = {Li, Yuhong and Huang, Yingbing and Yang, Bowen and Venkitesh, Bharat and Locatelli, Acyr and Ye, Hanchen and Cai, Tianle and Lewis, Patrick and Chen, Deming},
  title = {{{SnapKV}: {LLM} Knows What You are Looking for Before Generation}},
  booktitle = {Advances in Neural Information Processing Systems},
  volume = {37},
  pages = {22947--22970},
  year = {2024},
  url = {https://proceedings.neurips.cc/paper_files/paper/2024/hash/28ab418242603e0f7323e54185d19bde-Abstract-Conference.html},
  doi = {10.52202/079017-0722}
}

@article{milakov2018online,
  author = {Milakov, Maxim and Gimelshein, Natalia},
  title = {{Online Normalizer Calculation for Softmax}},
  journal = {arXiv preprint arXiv:1805.02867},
  year = {2018},
  url = {https://arxiv.org/abs/1805.02867},
  doi = {10.48550/arXiv.1805.02867}
}

@inproceedings{chuang2024lookback,
  title        = {{Lookback Lens: Detecting and Mitigating Contextual Hallucinations in Large Language Models Using Only Attention Maps}},
  author       = {Chuang, Yung-Sung and Qiu, Linlu and Hsieh, Cheng-Yu and Krishna, Ranjay and Kim, Yoon and Glass, James R.},
  booktitle    = {Proceedings of the 2024 Conference on Empirical Methods in Natural Language Processing},
  year         = {2024},
  pages        = {1419--1436},
  url          = {https://aclanthology.org/2024.emnlp-main.84/},
  doi          = {10.18653/v1/2024.emnlp-main.84}
}

@inproceedings{sriramanan2024llmcheck,
  author       = {Sriramanan, Gaurang and Bharti, Siddhant and Sadasivan, Vinu Sankar and Saha, Shoumik and Kattakinda, Priyatham and Feizi, Soheil},
  title        = {{LLM-Check: Investigating Detection of Hallucinations in Large Language Models}},
  booktitle    = {Advances in Neural Information Processing Systems},
  year         = {2024},
  volume       = {37},
  pages        = {34188--34216},
  url          = {https://proceedings.neurips.cc/paper_files/paper/2024/hash/3c1e1fdf305195cd620c118aaa9717ad-Abstract-Conference.html},
  doi          = {10.52202/079017-1077}
}

@inproceedings{bazarova2025toha,
  title        = {{Hallucination Detection in {LLM}s with Topological Divergence on Attention Graphs}},
  author       = {Bazarova, Alexandra and Volodichev, Andrei and Yugay, Aleksandr and Shulga, Andrey and Ermilova, Alina and Polev, Konstantin and Belikova, Julia and Parchiev, Rauf and Simakov, Dmitry and Savchenko, Maxim and Savchenko, Andrey and Barannikov, Serguei and Zaytsev, Alexey},
  booktitle    = {Proceedings of the 64th Annual Meeting of the {A}ssociation for {C}omputational {L}inguistics (Volume 1: Long Papers)},
  year         = {2026},
  pages        = {15449--15470},
  url          = {https://aclanthology.org/2026.acl-long.704/},
  doi          = {10.18653/v1/2026.acl-long.704}
}

@article{kuleshova2026intrygue,
  author       = {Kuleshova, Alexandra and Volodichev, Andrei and Kotova, Daria and Zaytsev, Alexey},
  title        = {{INTRYGUE: Induction-Aware Entropy Gating for Reliable RAG Uncertainty Estimation}},
  journal      = {arXiv preprint arXiv:2603.21607},
  year         = {2026},
  doi          = {10.48550/arXiv.2603.21607},
  note         = {Version 3, 31 August 2026. Accepted to EMNLP 2026 (Main)},
  url          = {https://arxiv.org/abs/2603.21607v3}
}

@inproceedings{niu2024ragtruth,
  title        = {{{RAGT}ruth: A Hallucination Corpus for Developing Trustworthy Retrieval-Augmented Language Models}},
  author       = {Niu, Cheng and Wu, Yuanhao and Zhu, Juno and Xu, Siliang and Shum, KaShun and Zhong, Randy and Song, Juntong and Zhang, Tong},
  booktitle    = {Proceedings of the 62nd Annual Meeting of the Association for Computational Linguistics (Volume 1: Long Papers)},
  year         = {2024},
  pages        = {10862--10878},
  url          = {https://aclanthology.org/2024.acl-long.585/},
  doi          = {10.18653/v1/2024.acl-long.585}
}

@article{ravi2024lynx,
  author       = {Ravi, Selvan Sunitha and Mielczarek, Bartosz and Kannappan, Anand and Kiela, Douwe and Qian, Rebecca},
  title        = {{Lynx: An Open Source Hallucination Evaluation Model}},
  journal      = {arXiv preprint arXiv:2407.08488},
  year         = {2024},
  url          = {https://arxiv.org/abs/2407.08488},
  doi          = {10.48550/arXiv.2407.08488}
}

@inproceedings{phantom2025,
  author       = {Ji, Lanlan and Seyler, Dominic and Kaur, Gunkirat and Hegde, Manjunath and Dasgupta, Koustuv and Xiang, Bing},
  title        = {{PHANTOM: A Benchmark for Hallucination Detection in Financial Long-Context QA}},
  booktitle    = {Advances in Neural Information Processing Systems},
  year         = {2025},
  volume       = {38},
  pages        = {140939--140968},
  url          = {https://papers.nips.cc/paper_files/paper/2025/hash/b8badadce3f482ba340ff870f4894441-Abstract-Datasets_and_Benchmarks_Track.html},
  doi          = {10.52202/085713-4235}
}

@inproceedings{tillet2019triton,
  author = {Tillet, Philippe and Kung, H. T. and Cox, David},
  title = {{Triton: An Intermediate Language and Compiler for Tiled Neural Network Computations}},
  booktitle = {Proceedings of the 3rd ACM SIGPLAN International Workshop on Machine Learning and Programming Languages},
  pages = {10--19},
  year = {2019},
  url = {https://dl.acm.org/doi/10.1145/3315508.3329973},
  doi = {10.1145/3315508.3329973}
}

@inproceedings{xiao2024streamingllm,
  author = {Xiao, Guangxuan and Tian, Yuandong and Chen, Beidi and Han, Song and Lewis, Mike},
  title = {{Efficient Streaming Language Models with Attention Sinks}},
  booktitle = {The Twelfth International Conference on Learning Representations},
  year = {2024},
  url = {https://openreview.net/forum?id=NG7sS51zVF}
}

@inproceedings{vazhentsev2025rauq,
  author       = {Vazhentsev, Artem and Rvanova, Lyudmila and Kuzmin, Gleb and Fadeeva, Ekaterina and Lazichny, Ivan and Panchenko, Alexander and Panov, Maxim and Sachan, Mrinmaya and Nakov, Preslav and Baldwin, Timothy and Shelmanov, Artem},
  title        = {{Efficient Hallucination Detection for {LLM}s Using Uncertainty-Aware Attention Heads}},
  booktitle    = {Proceedings of the 43rd International Conference on Machine Learning},
  series       = {Proceedings of Machine Learning Research},
  volume       = {306},
  pages        = {123596--123627},
  year         = {2026},
  url          = {https://proceedings.mlr.press/v306/vazhentsev26a.html}
}

@inproceedings{binkowski2026sinkprobe,
  author       = {Binkowski, Jakub and Adamczewski, Kamil and Kajdanowicz, Tomasz Jan},
  title        = {{Attention Sinks as Internal Signals for Hallucination Detection in Large Language Models}},
  booktitle    = {Proceedings of the 43rd International Conference on Machine Learning},
  series       = {Proceedings of Machine Learning Research},
  volume       = {306},
  pages        = {8302--8319},
  year         = {2026},
  url          = {https://proceedings.mlr.press/v306/binkowski26a.html}
}

@inproceedings{chen2026triviaplus,
  author       = {Chen, Wenbo and Padmanabhan, Veena and Giyahchi, Tootiya and Wong, Elaine and Akoglu, Leman},
  title        = {{Rethinking Evaluation for LLM Hallucination Detection: A Desiderata, A New RAG-based Benchmark, New Insights}},
  booktitle    = {Proceedings of the 64th Annual Meeting of the Association for Computational Linguistics (Volume 1: Long Papers)},
  year         = {2026},
  pages        = {14912--14931},
  url          = {https://aclanthology.org/2026.acl-long.680/},
  doi          = {10.18653/v1/2026.acl-long.680}
}

@inproceedings{han2024wildguard,
  title     = {{WildGuard}: Open One-stop Moderation Tools for Safety Risks, Jailbreaks, and Refusals of {LLMs}},
  author    = {Han, Seungju and Rao, Kavel and Ettinger, Allyson and Jiang, Liwei and Lin, Bill Yuchen and Lambert, Nathan and Choi, Yejin and Dziri, Nouha},
  booktitle = {Advances in Neural Information Processing Systems},
  volume    = {37},
  pages     = {8093--8131},
  year      = {2024},
  doi       = {10.52202/079017-0261},
  url       = {https://proceedings.neurips.cc/paper_files/paper/2024/hash/0f69b4b96a46f284b726fbd70f74fb3b-Abstract-Datasets_and_Benchmarks_Track.html}
}

@inproceedings{ghosh2025aegis2,
  title     = {{AEGIS2.0}: A Diverse {AI} Safety Dataset and Risks Taxonomy for Alignment of {LLM} Guardrails},
  author    = {Ghosh, Shaona and Varshney, Prasoon and Sreedhar, Makesh Narsimhan and Padmakumar, Aishwarya and Rebedea, Traian and Varghese, Jibin Rajan and Parisien, Christopher},
  booktitle = {Proceedings of the 2025 Conference of the Nations of the Americas Chapter of the Association for Computational Linguistics: Human Language Technologies (Volume 1: Long Papers)},
  pages     = {5992--6026},
  year      = {2025},
  doi       = {10.18653/v1/2025.naacl-long.306},
  url       = {https://aclanthology.org/2025.naacl-long.306/}
}

@inproceedings{chwang2024androids,
  author    = {{CH-Wang}, Sky and Van Durme, Benjamin and Eisner, Jason and Kedzie, Chris},
  title     = {{Do Androids Know They're Only Dreaming of Electric Sheep?}},
  booktitle = {Findings of the Association for Computational Linguistics: ACL 2024},
  pages     = {4401--4420},
  year      = {2024},
  url       = {https://aclanthology.org/2024.findings-acl.260/},
  doi       = {10.18653/v1/2024.findings-acl.260}
}

@inproceedings{zheng2023judging,
  author    = {Zheng, Lianmin and Chiang, Wei-Lin and Sheng, Ying and Zhuang, Siyuan and Wu, Zhanghao and Zhuang, Yonghao and Lin, Zi and Li, Zhuohan and Li, Dacheng and Xing, Eric P. and Zhang, Hao and Gonzalez, Joseph E. and Stoica, Ion},
  title     = {{Judging LLM-as-a-Judge with MT-Bench and Chatbot Arena}},
  booktitle = {Advances in Neural Information Processing Systems 36 (Datasets and Benchmarks Track)},
  year      = {2023},
  volume    = {36},
  pages     = {46595--46623},
  url       = {https://proceedings.neurips.cc/paper_files/paper/2023/hash/91f18a1287b398d378ef22505bf41832-Abstract-Datasets_and_Benchmarks.html},
  doi       = {10.52202/075280-2020}
}

@inproceedings{tang2024minicheck,
  author    = {Tang, Liyan and Laban, Philippe and Durrett, Greg},
  title     = {{MiniCheck}: Efficient Fact-Checking of {LLMs} on Grounding Documents},
  booktitle = {Proceedings of the 2024 Conference on Empirical Methods in Natural Language Processing},
  pages     = {8818--8847},
  year      = {2024},
  url       = {https://aclanthology.org/2024.emnlp-main.499/},
  doi       = {10.18653/v1/2024.emnlp-main.499}
}

@misc{li2024hhem,
  author    = {Li, Miaoran and Luo, Rogger and Mendelevitch, Ofer},
  title     = {{HHEM-2.1-Open}},
  year      = {2024},
  publisher = {Hugging Face},
  url       = {https://huggingface.co/vectara/hallucination_evaluation_model},
  doi       = {10.57967/hf/3240}
}

@misc{qwen2026qwen35,
  author    = {{Qwen Team}},
  title     = {{Qwen3.5-9B}},
  year      = {2026},
  publisher = {Hugging Face},
  url       = {https://huggingface.co/Qwen/Qwen3.5-9B}
}

@article{gemma2026gemma4,
  author  = {{Gemma Team}},
  title   = {{Gemma 4 Technical Report}},
  journal = {arXiv preprint arXiv:2607.02770},
  year    = {2026},
  doi     = {10.48550/arXiv.2607.02770},
  url     = {https://arxiv.org/abs/2607.02770}
}

@misc{ibm2026granite42,
  author    = {{IBM Granite Team}},
  title     = {{Granite-4.2-8B}},
  year      = {2026},
  publisher = {Hugging Face},
  url       = {https://huggingface.co/ibm-granite/granite-4.2-8b}
}

@article{liu2026ministral3,
  author  = {Liu, Alexander H. and Khandelwal, Kartik and Subramanian, Sandeep and others},
  title   = {{Ministral 3}},
  journal = {arXiv preprint arXiv:2601.08584},
  year    = {2026},
  doi     = {10.48550/arXiv.2601.08584},
  url     = {https://arxiv.org/abs/2601.08584}
}

@article{zuo2025falconh1,
  author  = {Zuo, Jingwei and Velikanov, Maksim and Chahed, Ilyas and others},
  title   = {{Falcon-H1: A Family of Hybrid-Head Language Models Redefining Efficiency and Performance}},
  journal = {arXiv preprint arXiv:2507.22448},
  year    = {2025},
  doi     = {10.48550/arXiv.2507.22448},
  url     = {https://arxiv.org/abs/2507.22448}
}

@article{dubey2024llama3,
  author  = {Dubey, Abhimanyu and Jauhri, Abhinav and Pandey, Abhinav and others},
  title   = {{The Llama 3 Herd of Models}},
  journal = {arXiv preprint arXiv:2407.21783},
  year    = {2024},
  doi     = {10.48550/arXiv.2407.21783},
  url     = {https://arxiv.org/abs/2407.21783}
}

@article{glm2024chatglm,
  author  = {{Team GLM}},
  title   = {{ChatGLM: A Family of Large Language Models from GLM-130B to GLM-4 All Tools}},
  journal = {arXiv preprint arXiv:2406.12793},
  year    = {2024},
  doi     = {10.48550/arXiv.2406.12793},
  url     = {https://arxiv.org/abs/2406.12793}
}

@article{abouelenin2025phi4mini,
  author  = {Abouelenin, Abdelrahman and Ashfaq, Atabak and Atkinson, Adam and others},
  title   = {{Phi-4-Mini Technical Report: Compact yet Powerful Multimodal Language Models via Mixture-of-LoRAs}},
  journal = {arXiv preprint arXiv:2503.01743},
  year    = {2025},
  doi     = {10.48550/arXiv.2503.01743},
  url     = {https://arxiv.org/abs/2503.01743}
}

@misc{cohere2024commandr7b,
  author    = {{Cohere Labs}},
  title     = {{Command R7B}},
  year      = {2024},
  publisher = {Hugging Face},
  url       = {https://huggingface.co/CohereLabs/c4ai-command-r7b-12-2024}
}

@article{yang2025qwen3,
  author  = {Yang, An and Li, Anfeng and Yang, Baosong and Zhang, Beichen and Hui, Binyuan and others},
  title   = {{Qwen3 Technical Report}},
  journal = {arXiv preprint arXiv:2505.09388},
  year    = {2025},
  doi     = {10.48550/arXiv.2505.09388},
  url     = {https://arxiv.org/abs/2505.09388}
}

@article{shrivastava2023llamas,
  author  = {Shrivastava, Vaishnavi and Liang, Percy and Kumar, Ananya},
  title   = {{Llamas Know What GPTs Don't Show: Surrogate Models for Confidence Estimation}},
  journal = {arXiv preprint arXiv:2311.08877},
  year    = {2023},
  url     = {https://arxiv.org/abs/2311.08877},
  doi     = {10.48550/arXiv.2311.08877}
}

@inproceedings{ulmer2024apricot,
  author    = {Ulmer, Dennis and Gubri, Martin and Lee, Hwaran and Yun, Sangdoo and Oh, Seong Joon},
  title     = {{Calibrating Large Language Models Using Their Generations Only}},
  booktitle = {Proceedings of the 62nd Annual Meeting of the Association for Computational Linguistics (Volume 1: Long Papers)},
  pages     = {15440--15459},
  year      = {2024},
  url       = {https://aclanthology.org/2024.acl-long.824/},
  doi       = {10.18653/v1/2024.acl-long.824}
}

@inproceedings{kapoor2024taught,
  author    = {Kapoor, Sanyam and Gruver, Nate and Roberts, Manley and Collins, Katherine and Pal, Arka and Bhatt, Umang and Weller, Adrian and Dooley, Samuel and Goldblum, Micah and Wilson, Andrew Gordon},
  title     = {{Large Language Models Must Be Taught to Know What They Don't Know}},
  booktitle = {Advances in Neural Information Processing Systems},
  volume    = {37},
  year      = {2024},
  url       = {https://proceedings.neurips.cc/paper_files/paper/2024/hash/9c20f16b05f5e5e70fa07e2a4364b80e-Abstract-Conference.html},
  doi       = {10.48550/arXiv.2406.08391}
}

@inproceedings{hayes2026pinocchio,
  author    = {Hayes, Kevin David and Pal, Arka and Zhang, Haosong and Goldstein, Tom and Goldblum, Micah},
  title     = {{Pinocchio: Fast Uncertainty Estimates for Black-Box Language Models}},
  booktitle = {Third Conference on Language Modeling},
  year      = {2026},
  url       = {https://openreview.net/forum?id=uPO7qVVOGN},
  doi       = {10.48550/arXiv.2609.24881}
}

@inproceedings{yeh2026lumina,
  author    = {Yeh, Samuel and Li, Sharon and Mallick, Tanwi},
  title     = {{LUMINA: Detecting Hallucinations in RAG System with Context--Knowledge Signals}},
  booktitle = {The Fourteenth International Conference on Learning Representations},
  year      = {2026},
  url       = {https://proceedings.iclr.cc/paper_files/paper/2026/file/f5846131aa6a72d1df3bd6d43a4a960b-Paper-Conference.pdf}
}

@article{singh2026proxyanalyzers,
  author  = {Singh, Akshita and Paudel, Prabesh and Roy, Siddhartha},
  title   = {{Hallucination Detection via Activations of Open-Weight Proxy Analyzers}},
  journal = {arXiv preprint arXiv:2605.07209},
  year    = {2026},
  url     = {https://arxiv.org/abs/2605.07209},
  doi     = {10.48550/arXiv.2605.07209}
}

@inproceedings{han2025factualityprobes,
  title     = {{Simple Factuality Probes Detect Hallucinations in Long-Form Natural Language Generation}},
  author    = {Han, Jiatong and Band, Neil and Razzak, Muhammed and Kossen, Jannik and Rudner, Tim G. J. and Gal, Yarin},
  booktitle = {Findings of the Association for Computational Linguistics: EMNLP 2025},
  pages     = {16209--16226},
  year      = {2025},
  url       = {https://aclanthology.org/2025.findings-emnlp.880/},
  doi       = {10.18653/v1/2025.findings-emnlp.880}
}

@inproceedings{azaria2023internal,
  title     = {{The Internal State of an LLM Knows When It's Lying}},
  author    = {Azaria, Amos and Mitchell, Tom},
  booktitle = {Findings of the Association for Computational Linguistics: EMNLP 2023},
  pages     = {967--976},
  year      = {2023},
  url       = {https://aclanthology.org/2023.findings-emnlp.68/},
  doi       = {10.18653/v1/2023.findings-emnlp.68}
}

@inproceedings{chen2024inside,
  author    = {Chen, Chao and Liu, Kai and Chen, Ze and Gu, Yi and Wu, Yue and Tao, Mingyuan and Fu, Zhihang and Ye, Jieping},
  title     = {{INSIDE: LLMs' Internal States Retain the Power of Hallucination Detection}},
  booktitle = {The Twelfth International Conference on Learning Representations},
  year      = {2024},
  pages     = {3056--3076},
  url       = {https://openreview.net/forum?id=Zj12nzlQbz}
}

@article{olsson2022induction,
  title   = {{In-context Learning and Induction Heads}},
  author  = {Olsson, Catherine and Elhage, Nelson and Nanda, Neel and Joseph, Nicholas and DasSarma, Nova and Henighan, Tom and Mann, Ben and Askell, Amanda and Bai, Yuntao and Chen, Anna and Conerly, Tom and Drain, Dawn and Ganguli, Deep and Hatfield-Dodds, Zac and Hernandez, Danny and Johnston, Scott and Jones, Andy and Kernion, Jackson and Lovitt, Liane and Ndousse, Kamal and Amodei, Dario and Brown, Tom and Clark, Jack and Kaplan, Jared and McCandlish, Sam and Olah, Chris},
  journal = {Transformer Circuits Thread},
  year    = {2022},
  url     = {https://transformer-circuits.pub/2022/in-context-learning-and-induction-heads/index.html},
  doi     = {10.48550/arXiv.2209.11895}
}

@inproceedings{gu2025attentionsink,
  author = {Gu, Xiangming and Pang, Tianyu and Du, Chao and Liu, Qian and Zhang, Fengzhuo and Du, Cunxiao and Wang, Ye and Lin, Min},
  title = {{When Attention Sink Emerges in Language Models: An Empirical View}},
  booktitle = {The Thirteenth International Conference on Learning Representations},
  year = {2025},
  url = {https://proceedings.iclr.cc/paper_files/paper/2025/hash/f1b04face60081b689ba740d39ea8f37-Abstract-Conference.html}
}

@inproceedings{chen2026toprovar,
  author = {Chen, Jiayu and Lin, Ruoyu and Zheng, Zihao and Li, Jingxin and Li, Maoliang and Luo, Guojie and Chen, Xiang},
  title = {{{ToProVAR}: Efficient Visual Autoregressive Modeling via Tri-Dimensional Entropy-Aware Semantic Analysis and Sparsity Optimization}},
  booktitle = {The Fourteenth International Conference on Learning Representations},
  year = {2026},
  url = {https://proceedings.iclr.cc/paper_files/paper/2026/file/8edb116d5b288b6a9bba4c16ab647702-Paper-Conference.pdf}
}

@inproceedings{liang2026vmonarch,
  author = {Liang, Cheng and Chen, Haoxian and Hou, Liang and Fan, Qi and Wu, Gangshan and Tao, Xin and Wang, Limin},
  title = {{{VMonarch}: Efficient Video Diffusion Transformers with Structured Attention}},
  booktitle = {Proceedings of the IEEE/CVF Conference on Computer Vision and Pattern Recognition},
  pages = {4613--4623},
  year = {2026},
  url = {https://openaccess.thecvf.com/content/CVPR2026/html/Liang_VMonarch_Efficient_Video_Diffusion_Transformers_with_Structured_Attention_CVPR_2026_paper.html}
}

@inproceedings{tu2025vlcache,
  author = {Tu, Dezhan and Vashchilenko, Danylo and Lu, Yuzhe and Xu, Panpan},
  title = {{{VL-Cache}: Sparsity and Modality-Aware {KV} Cache Compression for Vision-Language Model Inference Acceleration}},
  booktitle = {The Thirteenth International Conference on Learning Representations},
  year = {2025},
  url = {https://openreview.net/forum?id=HMrcv7Q4Ub}
}

\appendix

\section*{Guide to the appendices}
The appendices follow the order of the main text.
\begin{description}[leftmargin=0pt,itemsep=1pt,parsep=0pt,font=\normalfont\bfseries]
\item[Related work.] Appendix~\ref{sec:related}.
\item[Method (Sections~\ref{sec:background}--\ref{sec:method}).] Definitions, tiers, and users of every statistic the kernels compute (Appendix~\ref{app:stats}); proof of Lemma~\ref{lem:separable} and the accumulator's merge and stability (Appendix~\ref{app:separable}); pseudocode and kernel implementation (Appendix~\ref{app:impl}).
\item[Kernel benchmark (Section~\ref{sec:kernel-eval}).] Hardware, software, and timing protocol (Appendix~\ref{app:setup}); numerical accuracy, Q1 (Appendix~\ref{app:numerics}); cost per tier and length scaling, Q2 (Appendix~\ref{app:cost}); whole-model latency, Q3 and Figure~\ref{fig:preview} (Appendix~\ref{app:preview}); memory (Appendix~\ref{app:memory}); a per-layer comparison with dense attention on short inputs (Appendix~\ref{app:fusion}).
\item[Proxy detection (Section~\ref{sec:evaluation}).] Datasets, HaluBench filtering, and text-only baselines (Appendix~\ref{app:data}); feature recipes, logit features, training and significance tests, full AUROC results, and LLM judges (Appendix~\ref{app:detection}); extraction cost (Appendices~\ref{app:preview} and~\ref{app:memory}).
\item[Additional experiment.] Cross-domain transfer of attention detectors on HaluBench (Appendix~\ref{app:transfer}).
\item[Judge prompts.] The four prompt templates of the LLM judges (Appendix~\ref{app:prompts}).
\end{description}

\section{Related Work}
\label{sec:related}
Online softmax, its associative merge, and its fusion with other reductions such as top-$k$ are established \citep{milakov2018online,rabe2021selfattention,liu2024ringattention,feng2024attentionrnn}, and the separable row statistics of Section~\ref{sec:accumulator} follow directly from it; recovering a weight from the stored LSE is how the FlashAttention backward pass recomputes attention \citep{dao2024flashattention2}; and attending from a block of new queries to a cached prefix with a shifted causal mask is the key--value-cache append that serving systems and their kernels already support \citep{kwon2023pagedattention,agrawal2024sarathi,ye2025flashinfer}. Column sums of attention over past queries are the eviction scores of key--value-cache compression: H$_2$O sums over all past queries and SnapKV over an observation window \citep{zhang2023h2o,li2024snapkv}. Similar fused kernels appear as implementation details elsewhere: an online entropy accumulator in FlashAttention for visual generation \citep{chen2026toprovar,liang2026vmonarch}, and a second key-major pass for column sums over a short query window in VL-Cache \citep{tu2025vlcache}.

Several hallucination detectors read attention: Lookback Lens uses context-versus-answer attention ratios \citep{chuang2024lookback}, LLM-Check studies internal signals of an auxiliary model \citep{sriramanan2024llmcheck}, SinkProbe probes attention-sink scores \citep{binkowski2026sinkprobe}, TOHA uses graph structure in attention \citep{bazarova2025toha}, and INTRYGUE gates predictive entropy by the sink rates of induction heads \citep{kuleshova2026intrygue}. These methods motivate our feature families.

Outside hallucination detection, guard models such as WildGuard and Aegis2.0 moderate the responses of other models with 7--8B decoders \citep{han2024wildguard,ghosh2025aegis2}. Our proxies follow the same setup for contextual hallucinations and read their own attention rather than generating a verdict.

Taken together, efficient-attention work fuses reductions into the pass to produce the attention output or to rank keys for eviction, while attention-based detectors read their statistics from the materialized attention matrix. We connect the two: we map detector statistics to exact paths that avoid the matrix, measure their cost, and use them in small proxy detectors.

\FloatBarrier
\section{Statistics computed by the kernels}
\label{app:stats}
Table~\ref{tab:stats} defines every statistic the response-stage kernels compute and the detectors that use it. Notation follows Sections~\ref{sec:background} and~\ref{sec:method}: a response row $i\in\mathcal R$ sees the keys $\mathcal J_i$, which are the $P$ prompt keys and the response keys $P\le j\le i$ minus padding; $t_i=i-P+1$ is the number of response keys it sees, including itself, and $n_i=|\mathcal J_i|$. The kernels return per-row or per-column statistics for each head; the host turns them into the features of the last column in $O(R)$ or $O(L)$ time per head, and a detector recipe reduces these over response rows and selected heads (Section~\ref{sec:in-domain}). For the fixed sink we write $p_{is}$ for the per-row weight, $c_s=\sum_ip_{is}$ for its column sum, and $\bar c_s$ for the normalized sink rate. Qwen3 defines no beginning-of-sequence token; we call its first non-padded token, which opens the chat template, BOS.

\begin{table*}[t]
\centering
\footnotesize
\setlength{\tabcolsep}{4pt}
\renewcommand{\arraystretch}{1.25}
\begin{tabular}{@{}>{\raggedright\arraybackslash}p{0.14\textwidth}>{\raggedright\arraybackslash}p{0.22\textwidth}cc>{\raggedright\arraybackslash}p{0.19\textwidth}>{\raggedright\arraybackslash}p{0.22\textwidth}@{}}
\toprule
Statistic & Definition (per head) & Tier & Shape & Kernel computation & Derived feature and use\\
\midrule
Log-sum-exp & $\mathrm{LSE}_i=\log\sum_{j\in\mathcal J_i}e^{s_{ij}}$ & Base & $N_h\times R$ & $m_i+\log\ell_i$ after the last key tile, as in FlashAttention & none; normalizer reused by T2 and T3\\
Entropy & $H_i=-\sum_jp_{ij}\log p_{ij}$ & T1 & $N_h\times R$ & $\mathrm{LSE}_i-A_i/\ell_i$ with $\alpha=1$, $g=s$ & $H_i/\log n_i$.\newline Use: ours (A$_7$)\\
Effective support & $N_{\mathrm{eff},2,i}=\bigl(\sum_jp_{ij}^2\bigr)^{-1}$ & T1 & $N_h\times R$ & $\ell_i^2/A_i$ with $\alpha=2$, $g=1$ & Use: ours (A$_7$)\\
Maximum weight & $\max_jp_{ij}$ & T1 & $N_h\times R$ & $1/\ell_i$, no accumulator & Use: ours (A$_7$)\\
Prompt mass & $\mu^{\mathrm P}_i=\sum_{j<P}p_{ij}$ & T1 & $N_h\times R$ & $A_i/\ell_i$ with $\alpha=1$, $g=\Ind[j<P]$ & lookback ratio.\newline Use: Lookback Lens \citep{chuang2024lookback}\\
Response mass & $\mu^{\mathrm R}_i=\sum_{j\ge P}p_{ij}$ & T1 & $N_h\times R$ & $1-\mu^{\mathrm P}_i$, derived; no extra state & lookback ratio.\newline Use: Lookback Lens\\
Log-diagonal & $\log p_{ii}$ & T2 & $N_h\times R$ & $s_{ii}-\mathrm{LSE}_i$ & mean over response rows.\newline Use: LLM-Check \citep{sriramanan2024llmcheck}, adapted\\
Previous-token weight & $p_{i,i-1}$ & T2 & $N_h\times R$ & $\exp(s_{i,i-1}-\mathrm{LSE}_i)$ & Use: RAUQ \citep{vazhentsev2025rauq}\\
Fixed-sink weight & $p_{is}$ for a fixed key $s$, e.g.\ the first non-padded token (BOS) & T2 & $N_h\times S\times R$ & $\exp(s_{is}-\mathrm{LSE}_i)$; the host sums it into $c_s$ & BOS sink rate $\bar c_s$.\newline Use: ours (A$_7$)\\
Column sum & $c_j=\sum_{i\in\mathcal R,\,i\ge j}p_{ij}$ for every key $j$ & T3 & $N_h\times L$ & key-major rescoring with $\mathrm{LSE}$ & top-$k$ sink rate, A$_{64}$.\newline Use: SinkProbe \citep{binkowski2026sinkprobe}, adapted; INTRYGUE \citep{kuleshova2026intrygue}; ours (A$_{64}$)\\
\bottomrule
\end{tabular}
\caption{Statistics computed by the response-stage kernels. Tier: computation path (Section~\ref{sec:taxonomy}); Base: written by the statistics-off forward. Shape: kernel output per layer for one sequence, with $N_h$ query heads, $R$ response rows, $L$ keys, and $S$ fixed sink keys. $A_i$ is the T1 accumulator of Equation~\ref{eq:weighted-state} for the given $\alpha$ and $g$ after the last key tile, and $\ell_i$ the running sum of the online softmax. Derived features are defined in the text; ``adapted'': restricted to response rows.}
\label{tab:stats}
\end{table*}

\paragraph{Statistics only on response rows.}
Every statistic our detector reads is indexed by a response query row, and its column statistics sum over response rows only (Section~\ref{sec:in-domain}). We therefore run the prompt stage with any production kernel and compute statistics only in the response stage. For most published features this changes nothing: the lookback ratio of Lookback Lens and the previous-token weight of RAUQ are defined on generated tokens \citep{chuang2024lookback,vazhentsev2025rauq}. Two features change. LLM-Check averages $\log p_{ii}$ over all tokens \citep{sriramanan2024llmcheck}, and SinkProbe averages a key's weight over all later tokens \citep{binkowski2026sinkprobe}; we restrict both to response rows (Table~\ref{tab:stats}). On long contexts this keeps the prompt from dominating both averages, but it is a different feature, and we have not compared the two versions in detection.

\paragraph{Features built from the statistics.}
\emph{Normalized entropy} divides $H_i$ by $\log n_i$, the entropy of uniform attention over the visible keys, so rows at different positions are comparable. The \emph{lookback ratio} of Lookback Lens compares the mean weight per prompt key with the mean weight per response key \citep{chuang2024lookback}:
\begin{equation*}
\mathrm{LR}_i=\frac{\mu^{\mathrm P}_i/P}{\mu^{\mathrm P}_i/P+\mu^{\mathrm R}_i/t_i}.
\end{equation*}
The \emph{log-diagonal} feature is the per-head mean $\frac1R\sum_{i\in\mathcal R}\log p_{ii}$, with each term clamped at $\log\epsilon$. It is the attention score of LLM-Check \citep{sriramanan2024llmcheck}, the log-determinant of a head's attention matrix, which for a causal, hence lower-triangular, matrix is the sum of $\log p_{ii}$ over its rows. The \emph{previous-token} feature is $p_{i,i-1}$ on each response row, set to zero on the first response row, which has no previous response token; RAUQ reads it on selected heads \citep{vazhentsev2025rauq}. Attention sinks are keys that receive a large share of attention regardless of content \citep{xiao2024streamingllm}; \emph{sink rates} measure how much attention response rows put on a single key \citep{gu2025attentionsink}. Each column sum is divided by the number of response rows that see its key, $\bar c_j=c_j/\min(R,\,L-j)$, so $\bar c_j$ is the mean weight on key $j$. The BOS sink rate is $\bar c_s$ for the first non-padded key $s$, which T2 obtains without the full column sums when every sequence in the batch has the same padding; otherwise, and for any other sink definition, the kernels fall back to T3. The top-$k$ sink rate is the mean of the $k$ largest $\bar c_j$ ($k=1$ by default). SinkProbe uses the same per-key sink score, the mean weight a key receives from later tokens, and feeds the $k$ largest scores of every head to a probe \citep{binkowski2026sinkprobe}. INTRYGUE gates the entropy of the next-token distribution, which comes from the logits rather than the attention, by the top-1 sink rates of induction heads, computed over response rows as here \citep{kuleshova2026intrygue}.

\paragraph{Not computed by our kernels.}
Stored attention maps, the attention-graph divergence MTopDiv \citep{bazarova2025toha}, and row top-$k$ mass are computed from the materialized response block. Row top-$k$ mass needs $O(k)$ state per row, which we keep out of the pass, so T1 does not contain every bounded-state reduction. Threshold counts and column entropy fit T3 but are not implemented.

\FloatBarrier
\section{Separable row reductions}
\label{app:separable}

\paragraph{Proof of Lemma~\ref{lem:separable}.}
By induction over key tiles, Equation~\ref{eq:update} keeps $A_u$ equal to Equation~\ref{eq:weighted-state}. After the last tile, $\ell=e^{\mathrm{LSE}-m}$, so
\[
\frac{A_u}{\ell^{\alpha_u}}
=\sum_j\left(\frac{e^{s_j-m}}{\ell}\right)^{\alpha_u}g_u(i,j,s_j)
=F_i^{(u)}.
\]
Thus each $F_i^{(u)}$, and any fixed function of these values and the LSE, is available after the pass with the stated state and work.\hfill$\square$

\paragraph{Merge and stability.}
Two accumulator states over disjoint key sets merge as
\[
\begin{aligned}
A_u={}&e^{\alpha_u(m_1-m)}A_{u,1}\\
&+e^{\alpha_u(m_2-m)}A_{u,2},\\
m={}&\max(m_1,m_2),
\end{aligned}
\]
an associative operation that also permits keys to be split across kernel instances or devices \citep{liu2024ringattention,feng2024attentionrnn,ye2025flashinfer}. Since $m'\ge m$ and $m'\ge s_j$, the rescaling factors in Equation~\ref{eq:update} lie in $(0,1]$, so no exponential overflows. The maximizing key contributes one, giving $1\le\ell\le|\mathcal J_i|$ and $|A_u|\le|\mathcal J_i|\max_j|g_u(i,j,s_j)|$.

\FloatBarrier
\section{Implementation}
\label{app:impl}

\paragraph{Statistics in the pass.}
\label{sec:in-pass}
The response-stage kernel applies the update of Equation~\ref{eq:update} inside its loop, next to the FlashAttention update (Figure~\ref{fig:tiled-statistics}b). A response query tile stays resident while key/value tiles stream across it, and each score tile updates both the FlashAttention state and the statistic accumulators before the next tile is loaded; nothing is retained per key. Algorithm~\ref{alg:fused} gives the pseudocode for all three kernels.

\begin{algorithm}[t]
\caption{Attention with statistics: pseudocode for Figure~\ref{fig:tiled-statistics}b--c. Black lines are the unchanged FlashAttention-2 forward (blue in Figure~\ref{fig:tiled-statistics}); \Stat{green} lines are our additions, tagged with their tier. Each kernel instance handles one query tile $I$ (or, in T3, one key tile $J$) of one head; $m,\ell,A_u$ hold one entry per query row. A key tile is visible to a query tile if some row of $I$ may attend to some key of $J$; the mask removes padded keys and, on tiles that cross the diagonal, future keys. $G_u=g_u(J,S)$, and $\mathcal S$ is an optional set of fixed sink columns, whose weights $p_{is}$ the host sums over rows into $c_s$. The LSE is part of the unchanged forward and belongs to no tier.}
\label{alg:fused}
\footnotesize
\begin{algorithmic}[1]
\State $O_{<P}\gets\mathrm{CausalAttn}(Q_{<P},K_{<P},V_{<P})$ \Comment{any kernel}
\For{each response query tile $I\subseteq\mathcal R$}
  \State $m\gets-\infty,\ \ell\gets0,\ O\gets0$
  \State \Stat{$A_u\gets0$ for $u=1,\dots,N_s$} \Tier{T1}
  \For{each key tile $J$ visible to $I$}
    \State $S\gets\tau Q_IK_J^\top$; mask
    \State $m'\gets\max(m,\mathrm{rowmax}(S))$
    \State $a\gets e^{m-m'};\ \tilde P\gets e^{S-m'}$
    \State \Stat{$A_u\gets a^{\alpha_u}A_u+\mathrm{rowsum}(\tilde P^{\alpha_u}\odot G_u)$}
    \State $O\gets aO+\tilde PV_J;\ \ell\gets a\ell+\mathrm{rowsum}(\tilde P)$
    \State $m\gets m'$
  \EndFor
  \State write $O/\ell$ and $\mathrm{LSE}\gets m+\log\ell$
  \State \Stat{write $F^{(u)}$ from $\mathrm{LSE},\ell,A_u/\ell^{\alpha_u}$} \Tier{T1}
\EndFor
\For{\Stat{each response row $i\in\mathcal R$}}
  \State \Stat{$\log p_{ii}\gets s_{ii}-\mathrm{LSE}_i$} \Tier{T2}
  \State \Stat{$p_{i,i-1}\gets\exp(s_{i,i-1}-\mathrm{LSE}_i)$}
  \State \Stat{$p_{is}\gets\exp(s_{is}-\mathrm{LSE}_i)$ for $s\in\mathcal S$}
\EndFor
\For{\Stat{each key tile $J$}}
  \State \Stat{$c_J\gets0$} \Tier{T3}
  \For{\Stat{each response query tile $I$ that sees $J$}}
    \State \Stat{$W\gets\exp(\tau Q_IK_J^\top-\mathrm{LSE}_I)$; mask}
    \State \Stat{$c_J\gets c_J+\mathrm{colsum}(W)$}
  \EndFor
  \State \Stat{write $c_J$}
\EndFor
\end{algorithmic}
\end{algorithm}

Entropy uses the unshifted score $g=s$ rather than $s-m$, since $H_i=\mathrm{LSE}_i-\sum_jp_{ij}s_{ij}$. Nonlinearity alone is not the boundary of T1: $\sum_jp_{ij}^2$ remains separable. The maximum weight $1/\ell_i$ requires $m$ to be the exact running maximum; a kernel that defers rescaling when the maximum grows would have to track it separately. The state is $N_s$ numbers per row for $N_s$ scalar accumulators, independent of $L$, and writing them for $N_h$ heads costs $O(N_hRN_s)$ memory traffic.

\paragraph{Kernels and attention backend.}
\label{sec:implementation}
A Hugging Face attention backend, installed only on the scored layers, splits each layer call into the two stages. The prompt stage calls the model's original attention backend (in our runs, PyTorch SDPA with the FlashAttention-2 or cuDNN backend). The response stage is a Triton kernel \citep{tillet2019triton} derived from the FlashAttention-2 causal forward, with $R$ query rows, $L$ keys, and the causal mask shifted by $P$; it accumulates the T1 statistics and writes the LSE. T2 is a separate kernel that reads $Q$, $K$, and the LSE. It does not depend on which T1 statistics are compiled into the response-stage kernel, it works after any forward kernel that writes the LSE, and its positions are runtime arguments, so any set of sink columns can be read without recompiling. T3 is a key-major kernel modeled on the FlashAttention-2 backward: column sums are its value-gradient step, $dV=\mathbf W^\top dO$ for the attention weights $\mathbf W$ and the output gradient $dO$, with $dO$ replaced by a column of ones on response rows. Each kernel instance owns one key tile, visits the response query tiles that see it, and writes its column sums once, without atomic additions. All kernels support grouped-query attention, bf16/fp16/fp8-e5m2 inputs, head dimensions 64/128/256, and a key keep mask; statistic families are compile-time flags. The kernels do not support sliding-window or soft-capped attention, both elementwise changes to the mask and the scores; layers that use them run the model's default attention without statistics.

\FloatBarrier
\section{Kernel benchmark setup}
\label{app:setup}
All timed runs use a single NVIDIA H100 NVL (Hopper, 94~GB, power-capped at 270~W; CUDA 12.9, PyTorch 2.11, Triton 3.6, cuDNN 9.17, Transformers 5.5) in bf16. The production baselines are PyTorch SDPA with the flash backend (FlashAttention-2) and the cuDNN backend; FlashAttention-3 is not installed, and cuDNN is the Hopper-specific baseline. Besides the Qwen3-8B shapes, $d=64$ uses the same head counts and $d=256$ uses 8/4 heads (Gemma-3 global layers). Kernel tiles are autotuned once per head dimension at $L=32\mathrm{K}$, $r=0.3$ and reused for every shape. Timings are medians over 100 repetitions after 25 warm-up runs with an L2 flush (configuration sweep), 15--80 repetitions depending on length (length scaling), or 20 repetitions after 3 warm-ups (per-layer comparison, Appendix~\ref{app:fusion}). Appendix~\ref{app:memory} describes how we measure memory. Sink positions are resolved outside the timed region. The host is shared, but no other process ran on the measured GPU.

\paragraph{Dense baseline.}
Every comparison with materialized attention uses the path of detectors that read \texttt{output\_attentions}: the attention weights of the response rows are computed as an fp32 softmax cast to bf16 and reduced in PyTorch.

\FloatBarrier
\section{Numerical accuracy (Q1)}
\label{app:numerics}
Unit tests compare the kernels against a dense PyTorch reference across dtypes (including fp8-e5m2), head dimensions, GQA ratios, padding masks, prompt/response splits, and statistic sets. Against an fp64 reference computed from the materialized matrix (row-chunked at 32K), we measure the absolute error on response rows of (i) the dense bf16 path (Appendix~\ref{app:setup}) and (ii) our kernels (T1 in the pass, T2 lookups, T3 column sums over the response rows), on random Gaussian $Q,K,V$ (response fraction 0.3) and on real $Q,K,V$ from four Qwen3-8B layers (2, 16, 24, 34) on the longest input of the PHANTOM 20K set \citep{phantom2025}, its prompt truncated so that the input has 2K, 8K, or 32K tokens, with a response of at most 156 tokens. Random inputs give near-uniform attention and hide sinks and peaked heads; the real inputs do not.

\begin{table}[t]
\centering
\footnotesize
\setlength{\tabcolsep}{3pt}
\begin{tabular}{lrrr}
\toprule
Statistic & dense & fused & ratio\\
\midrule
Entropy $H$ (T1) & \sci{1.6}{-2} & \sci{3.5}{-6} & \sci{4.5}{3}\\
Effective support $N_{\mathrm{eff},2}$ (T1) & \sci{5.3}{0} & \sci{5.9}{-3} & \sci{9.0}{2}\\
Maximum weight (T1) & \sci{4.3}{-3} & \sci{1.5}{-7} & \sci{2.9}{4}\\
Prompt mass (T1) & \sci{2.2}{-3} & \sci{1.3}{-7} & \sci{1.7}{4}\\
LSE (Base) & \sci{1.1}{-2} & \sci{1.1}{-6} & \sci{1.0}{4}\\
$\log p_{ii}$ (T2) & \sci{2.0}{-2} & \sci{1.2}{-6} & \sci{1.7}{4}\\
$p_{i,i-1}$ (T2) & \sci{5.0}{-4} & \sci{3.3}{-8} & \sci{1.5}{4}\\
BOS column sum $c_s$ (T2) & \sci{5.3}{-2} & \sci{7.1}{-5} & \sci{7.4}{2}\\
Column sum (T3) & \sci{2.2}{-4} & \sci{6.6}{-8} & \sci{3.3}{3}\\
Attention output & \sci{3.5}{-3} & \sci{4.8}{-4} & 7.2\\
\bottomrule
\end{tabular}
\caption{Mean absolute error on response rows against an fp64 reference for materialized attention (the dense bf16 path) and our bf16 kernels, and their ratio. Each entry averages the per-head error over heads and layers and reports the worst input type (random or PHANTOM) and length (2K, 8K, 32K). Our statistics are about three to four orders of magnitude closer to fp64; they have the lower error for every statistic, input, and length, in both mean and maximum error.}
\label{tab:numerics}
\end{table}

The dense path rounds each probability to bf16 (8 bits of mantissa) before reducing, while the kernels reduce fp32 probabilities that never leave registers. The T2 lookups and T3 column sums reuse the stored LSE, so moving a statistic out of the pass costs time, not accuracy. Both paths rank heads by their per-head means over response rows as fp64 does, with Spearman correlation 1.000 for our kernels and at least 0.998 for the dense path. The entropy accumulator uses the unshifted score $g=s$ (Appendix~\ref{app:impl}); its per-head mean error stays below \sci{4.7}{-5} even for heads with $H<0.25$ at 32K, where the dense error reaches \sci{1.4}{-2}.

\FloatBarrier
\section{Cost per tier and length scaling (Q2)}
\label{app:cost}
\begin{table*}[t]
\centering
\footnotesize
\setlength{\tabcolsep}{2.8pt}
\newcommand{\nr}[2]{$-#1$\,--\,$#2$}
\begin{tabular}{lrrrrrrrrr}
\toprule
& \multicolumn{3}{c}{$d=64$} & \multicolumn{3}{c}{$d=128$} & \multicolumn{3}{c}{$d=256$}\\
\cmidrule(lr){2-4}\cmidrule(lr){5-7}\cmidrule(lr){8-10}
Overhead (\%) & $r{=}.1$ & $r{=}.3$ & $r{=}.5$ & $r{=}.1$ & $r{=}.3$ & $r{=}.5$ & $r{=}.1$ & $r{=}.3$ & $r{=}.5$\\
\midrule
\cfg{1} (in pass) vs.\ \base{} & 9--12 & 22 & 27--28 & 3--7 & 12--15 & 9 & 1--4 & 6--9 & 2--6\\
\cfg{2} (+ lookups) vs.\ \base{} & 9--12 & 21--22 & 27--30 & 3--6 & 12--20 & 9--10 & 1--4 & 6--10 & 4--7\\
\cfg{3} (+ col.\ sums) vs.\ \base{} & 25--26 & 64 & 87 & 16--19 & 39--44 & 53--58 & 19--26 & 39--43 & 54--57\\
\midrule
\base{} vs.\ full cuDNN & 6--11 & 12--19 & 16--20 & 4--9 & 7--15 & 10--17 & 10--13 & 20--25 & 28--34\\
\base{} vs.\ full FA2 & 0--3 & \nr{4}{-1} & \nr{9}{-8} & \nr{2}{+1} & \nr{8}{-5} & \nr{13}{-12} & \nr{5}{-2} & \nr{16}{-11} & \nr{21}{-20}\\
\cfg{2} vs.\ full cuDNN & 15--25 & 36--45 & 48--55 & 7--16 & 19--38 & 20--29 & 14 & 28--37 & 33--43\\
\bottomrule
\end{tabular}
\caption{The numbers behind Figure~\ref{fig:tier-cost}a (Q2): latency overhead (\%) of one attention layer, as a range over $L\in\{16\mathrm{K},32\mathrm{K}\}$. Upper rows: each configuration of Section~\ref{sec:kernel-eval} over the statistics-off split \base{}, with the prompt stage on SDPA cuDNN. Lower rows: \base{} over SDPA on the whole sequence, with the prompt stage on the same backend, and the cost of all row statistics (\cfg{2}) over full SDPA cuDNN, the fastest production kernel here. H100 NVL, bf16.}
\label{tab:tiers-cost}
\end{table*}

\paragraph{Notes on Table~\ref{tab:tiers-cost}.}
\base{} runs the response rows on our kernel, which is faster than FlashAttention-2 on Hopper and slower than cuDNN, so as $r$ grows, \base{} gains on full FA2 and loses to full cuDNN. On the response stage alone, \cfg{3} takes 1.7--1.8$\times$ the time of \base{} at $d\ge128$ and 2.1--2.2$\times$ at $d=64$; the key-major T3 kernel does not spill registers, so this cost is the arithmetic of a second $QK^\top$ pass over the response block, not register pressure. Inputs of 4K tokens are excluded: a layer then takes 0.3--1.0~ms, and its ratios reflect the fixed host time per call (Appendix~\ref{app:preview}).

\paragraph{Length scaling.}
Figure~\ref{fig:tier-cost}b--c times the same layer at 20 geometrically spaced lengths from 1K to 100K tokens, at $d=128$ (32 query heads, 8 KV heads, batch 1) and $r\in\{0.1,0.3\}$: full FlashAttention-2 and cuDNN without statistics, and the split layers \base{}, \cfg{2} (with a fixed BOS sink), and \cfg{3}, with the prompt on cuDNN. From 15K to 100K tokens, each overhead stays within a band that does not grow with $L$. At $r=0.3$, \cfg{2} adds 3--14\% over \base{}, \cfg{3} adds 32--47\%, and \base{} takes 4--14\% longer than cuDNN; at $r=0.1$, the same overheads are $-2$ to $+13\%$, 9--22\%, and $-2$ to $+8\%$. Below about 4K tokens, the split layer stays at 0.3--0.5~ms down to 1K tokens, against 0.04--0.07~ms for SDPA; this floor is host time per call (Appendix~\ref{app:preview}).

\FloatBarrier
\section{Whole-model latency and extraction cost (Q3)}
\label{app:preview}
\paragraph{Setup of Figure~\ref{fig:preview}.}
Figure~\ref{fig:preview} and the figures of this appendix time one forward of Qwen3-8B (all 32 heads of all 36 layers) and of Qwen3-0.6B (all 16 heads of all 28 layers) on PHANTOM and TRIVIA+ inputs. Our kernels compute the statistics underlying Lookback Lens, LLM-Check, RAUQ, INTRYGUE, and SinkProbe; the baseline is the same forward without features. Attention outside our kernels runs on SDPA with FlashAttention-2 for prompts up to 8K tokens and with its default cuDNN backend for longer ones, the same in both forwards; Q3 uses FlashAttention-2 throughout. Means are taken per length bin, and bins with fewer than 12 samples are omitted. Batch 1, H100 NVL, bf16.

\paragraph{Data for Q3.}
From RAGTruth QA, Q3 takes the 989 responses of Llama-2-70B-chat, median 0.5K and at most 1.0K tokens. From PHANTOM, it takes the split of the 10-K filings set that PHANTOM labels 2,000 tokens, with the answer-bearing passage in the middle: 980 samples with median 2.6K and at most 8.4K Qwen3 tokens. Token counts include the prompt template.

\paragraph{Dense baseline for Q3.}
Dense (Appendix~\ref{app:setup}) takes each layer's output from SDPA, then recomputes the attention weights of the response rows for all 32 heads at once, reduces them, and frees them before the next layer. Keeping every layer's full weights from the forward, as \texttt{output\_attentions} does, would take 152~GiB at 8.4K tokens, more than the GPU holds; Dense instead trades one recomputation of $QK^\top$ and the softmax per layer for the memory of one layer's response block, $32RL$ weights. Taking the response output from the materialized weights instead would avoid scoring the response rows twice but add a product with $V$.

\begin{figure*}[t]
\centering
\includegraphics[width=\textwidth]{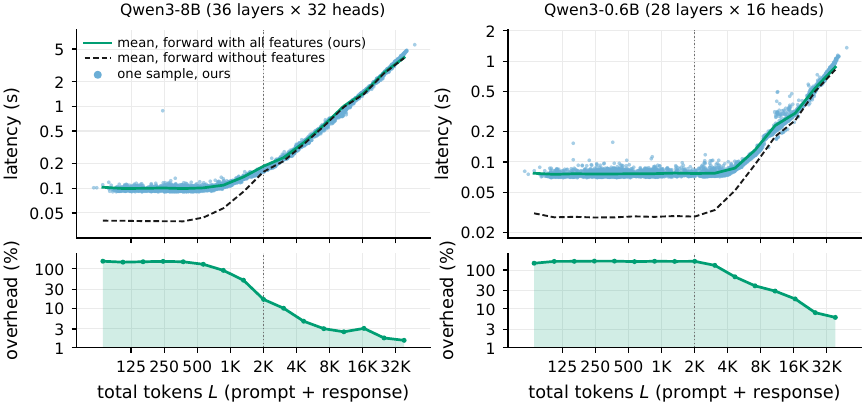}
\caption{Figure~\ref{fig:preview} at every length, for Qwen3-8B (left) and Qwen3-0.6B (right, all 16 heads of all 28 layers), with the overhead on a log scale. Dotted line: the 2K cut of Figure~\ref{fig:preview}. Same inputs, backends, and binning; batch 1, H100 NVL, bf16.}
\label{fig:preview-full}
\end{figure*}

\begin{table}[t]
\centering
\footnotesize
\begin{tabular}{lrr}
\toprule
Per forward & Floor & Ours\\
\midrule
Latency (ms) & 33 & 108\\
GPU busy (ms) & 13 & 18\\
GPU kernels & 2{,}005 & 5{,}173\\
Host--device syncs & 1 & 361\\
\bottomrule
\end{tabular}
\caption{Where the time of a short forward goes: Qwen3-8B on five TRIVIA+ inputs of 90--253 tokens, all 36 layers scored, FlashAttention-2, H100 NVL. GPU busy is the summed device time of all kernels (PyTorch profiler); syncs are counted with PyTorch's synchronization debug mode.}
\label{tab:short-profile}
\end{table}

\paragraph{Fixed host cost on short inputs.}
Figure~\ref{fig:preview-full} repeats Figure~\ref{fig:preview} without the 2K cut, with one panel per model. Below about 600 tokens for Qwen3-8B and 3K for Qwen3-0.6B, a forward with features takes the same time at every length: the features add a fixed 60 and 47~ms, about 1.7~ms per scored layer for both models, although a Qwen3-0.6B layer does a fraction of the work of a Qwen3-8B layer.

This cost is host time, not GPU work (Table~\ref{tab:short-profile}). For every scored layer, the wrapper runs Python code, launches 88 more kernels than the plain layer, most of them small element-wise operations that turn row statistics and column sums into features, and reads ten values back from the GPU: the prompt length, response length, and padding of the sample, twice; a check for padding in the prompt block; and three index lists for the selected heads. Each read stalls the host until the GPU queue drains. Under the Python profiler, launching our three kernels takes 0.8~ms per layer, assembling features another 0.8~ms, and copying the configuration of every scored layer to switch its attention backend 9~ms per forward. None of this grows with the input. The ten values could be computed once per forward, and capturing the scored layers in a CUDA graph would remove the launch cost. A split layer called on its own, as in Appendices~\ref{app:cost} and~\ref{app:fusion}, pays only its setup and the 19 kernel launches for all row statistics, about 0.3--0.6~ms of host time per call.

\FloatBarrier
\section{Memory}
\label{app:memory}
\begin{figure}[t]
\centering
\includegraphics[width=\columnwidth]{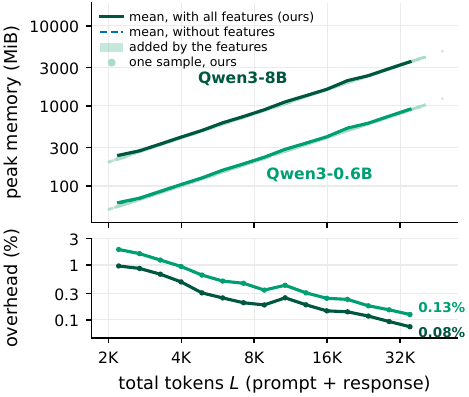}
\caption{Figure~\ref{fig:preview} for peak memory: the peak of each forward above the memory allocated before it, so the model weights are excluded, with all features (solid, points) and without (dashed). Same runs, inputs of at least 2K tokens, and binning; batch 1, H100 NVL, bf16.}
\label{fig:preview-memory}
\end{figure}

\paragraph{Measurement.}
All memory numbers count PyTorch allocations: \texttt{max\_memory\_allocated} after \texttt{reset\_peak\_memory\_stats}, minus the memory allocated before the measured call. They exclude memory that the caching allocator has reserved but not handed out, so \texttt{nvidia-smi} shows more. A buffer counts only if it is allocated at the moment of the peak. For one layer (Q2 and Appendix~\ref{app:fusion}), $Q$, $K$, and $V$ are allocated before the call, so the extra peak is what the layer allocates: its output, the statistics, and temporary buffers. For a whole forward (Q3 and Figure~\ref{fig:preview-memory}), the weights and inputs are allocated before the call, so the extra peak covers activations and features but not the weights.

\paragraph{One attention layer (Q2).}
At 32K and $d=128$, the extra peak of a layer is 520--532~MiB in every configuration, against 256~MiB for SDPA on the whole sequence, and 1.6 against 0.8~GiB at 100K. Our kernel allocates the statistic outputs even with statistics off, so \base{} has the same peak; the statistics take 7.6--20~MiB of it, 1.5--4\%. Most of the difference from SDPA is the concatenation of the two stage outputs: the layer must return all $L$ rows, so it copies them into one output-sized tensor while both are still allocated. The attention matrix of the same layer would take 64~GiB at 32K.

\paragraph{Whole forward (Q3).}
The concatenated copy is freed within its layer and does not reach the peak of a whole forward. Over the floor, our extra peak grows by 26--32~KiB per response token and not with the prompt (least-squares fit on each dataset), whereas an output-sized copy at the peak would add 8~KiB per token of the sequence. The excess is mostly the extracted statistics, 22.5~KiB per response token for Qwen3-8B, which stay on the GPU until the forward returns. Dense holds the response block of one layer, $32RL$ weights, at a time (Appendix~\ref{app:preview}). Ours adds at most 3.2\% to the floor on PHANTOM, whose prompts all exceed 1.1K tokens, and 6\% at the median and up to 17\% on the short inputs of RAGTruth QA.

\paragraph{Longer inputs.}
Figure~\ref{fig:preview-memory} extends the whole-forward measurement to 35K tokens, on the runs of Figure~\ref{fig:preview} (Appendix~\ref{app:preview}). Without features, a forward peaks at 104.5~KiB per token above the weights for Qwen3-8B and 26.6~KiB for Qwen3-0.6B. Our features store 20~bytes per head, layer, and response token, the five per-token features in fp32 (22.5 and 8.75~KiB per response token), plus 1.0 and 0.5~MiB that the wrapper holds between forwards; the measured peak grows by about 28 and 11~KiB per response token (least-squares fit over these runs). On short inputs this is a larger share of a smaller forward, up to 14\% for Qwen3-8B below 100 tokens. Below about 1K tokens, the peaks of Qwen3-0.6B jump between levels at multiples of 128 tokens in both forwards. This matches the scratch buffer that FlashAttention-2 sizes by the number of query tokens when it splits a short input across more thread blocks: the plain forward attends with all $L$ queries and ours runs the prompt with $P$, so near a step the two can land on different levels, by up to 31~MiB, independently of the features.

\paragraph{Reducing memory further.}
The concatenation is avoidable. If the prompt goes through a separate prefill, or its key--value cache is reused from generation, the response stage reads the cached keys and values and returns only its $R$ rows, so nothing is concatenated; the price is that the prompt cache stays allocated across all layers. The per-token features stay on the GPU until the forward returns; reducing them to per-head aggregates within each layer would make their memory independent of the response length. Finally, we compare with dense end to end only up to 8.4K tokens, where a dense pipeline still fits; a long-context end-to-end comparison is future work.

\FloatBarrier
\section{Per-layer comparison on short inputs}
\label{app:fusion}
Q3 compares whole forwards; here we time a single attention layer with statistics, ours against dense materialization, on the short real inputs of Figure~\ref{fig:q3-scaling}, and check that both give the same features.

We use the 989 RAGTruth QA inputs of Q3 (Appendix~\ref{app:preview}; response fraction $\approx$0.3), on which every method, including dense materialization, fits in memory. We capture $Q,K,V$ of 48 samples (16 per length tercile) from 11 Qwen3-8B layers between layers 3 and 24 and time each method per layer call, for all 32 heads, at batch 1 and at batch 8 with padding and a keep mask. \emph{Dense} materializes the attention matrix for all heads at once (Appendix~\ref{app:setup}); \emph{Dense-loop} does the same one head at a time, as the dense fallback of our pipeline does by default; \emph{Ours} is our split layer (prompt rows on SDPA, response rows on our kernels). Every method obtains the attention output from the same PyTorch SDPA call and computes, on response rows only, the T1 and T2 statistics (entropy group, prompt mass, BOS sink, diagonal, previous token) and the full T3 column sums, from which the sink values of A$_{64}$ are read.

\paragraph{Latency and memory.} At these lengths both methods are bound by host time, not GPU work: profiling a 1K-token layer with all row statistics shows 0.62~ms of host time to enqueue 19 kernels that together run for 0.10~ms on the GPU, against 0.06~ms for one cuDNN call. Dense is therefore slightly faster than ours at batch 1, with a median of 1.19 against 1.26~ms per layer call (ours/Dense $=1.04$--$1.09$ in every length tercile), and batching amortizes the host time (Appendix~\ref{app:preview}), so ours is faster at batch 8, 4.71 against 7.60~ms. Dense-loop takes 35.9~ms at batch 1; most of this gap is its loop over heads, not materialization, so we use batched Dense as the dense baseline. The extra peak memory (Appendix~\ref{app:memory}) of a layer call is 14~MiB for ours against 116~MiB for Dense at batch 1, and 136~MiB against 1.4~GiB at batch 8. Appendix~\ref{app:cost} gives the dependence on $L$.

\paragraph{Feature agreement.} Ours differs from Dense by a mean relative 0.04\% (the log-diagonal feature of LLM-Check, Appendix~\ref{app:stats}) to 1.6\% (previous-token weight). The largest absolute difference, 6.8, is on effective support (typical values $\approx12$); against fp64, the mean error of effective support is 5.3 for Dense and \sci{5.9}{-3} for ours (Table~\ref{tab:numerics}).

\FloatBarrier
\section{Data}
\label{app:data}
Table~\ref{tab:det-data} gives the splits of the three benchmarks of Section~\ref{sec:evaluation}.

\textbf{RAGTruth QA} \citep{niu2024ragtruth}: the question-answering subset, with the answers of quality \emph{good} from all six generators. We keep the official test split and divide the training answers 80/20 by query into training and validation (seed 42); groups are queries.

\textbf{HaluBench} \citep{ravi2024lynx}: 14,900 answers from six sources: HaluEval (10,000); DROP, CovidQA, PubMedQA, and FinanceBench (1,000 each); and RAGTruth (900). We filter and split it ourselves (below); the transfer experiment of Appendix~\ref{app:transfer} uses the same split.

\textbf{TRIVIA+} \citep{chen2026triviaplus}: the original training, validation, and test splits; groups are normalized source articles.

Within each dataset, every feature set and proxy uses the same training, validation, and test examples, and groups never cross splits. The datasets may overlap with each other, which does not matter here because no detector is applied across datasets.

\begin{table}[h]
\centering
\footnotesize
\setlength{\tabcolsep}{4pt}
\begin{tabular}{lrrrrr}
\toprule
Dataset & Train & Val & Test & Hal. & Grp.\\
\midrule
RAGTruth QA & 3921 & 971 & 875 & 160 & 150\\
HaluBench & 4206 & 898 & 904 & 409 & 546\\
TRIVIA+ & 2263 & 316 & 645 & 224 & 358\\
\bottomrule
\end{tabular}
\caption{Splits used in Section~\ref{sec:evaluation}. Hal.: hallucinated test answers; Grp.: test groups for the bootstrap.}
\label{tab:det-data}
\end{table}

\paragraph{HaluBench filtering and splitting.}
In HaluEval, faithful answers are mostly short entity names and hallucinated ones long sentences, so the answer length in words alone reaches an AUROC of 0.970 on its 10,000 answers and 0.720 on all 14,900. We therefore subsample HaluEval to equally many faithful and hallucinated answers at every answer length in words, which keeps 1,108 of its answers and 6,008 in total; after this, length gives an AUROC of 0.500 on HaluEval and 0.586 overall, where shorter answers are slightly more often hallucinated. We then group near-duplicate contexts within each source: two contexts are linked if at least 30\% of the word 5-grams of the shorter one occur in the other, and two FinanceBench filings also if their word sets have a Jaccard similarity of at least 0.6 or they share at least six numbers of four or more digits other than years. Groups are the connected components of these links; whole groups are assigned to splits to approach 70/15/15\% within every source and label (seed 42), and they are the units of the bootstrap.

\paragraph{Text-only baselines.}
To measure how much of each label the answer text reveals without the context, we score the test answers by their length and by a TF--IDF classifier on the answer alone (word unigrams and bigrams, the logistic regression of Section~\ref{sec:selection}, with regularization and threshold chosen on validation); Table~\ref{tab:text-baselines} gives the results. On unfiltered HaluBench, TF--IDF on the answer alone reaches 0.842 AUROC: much of the label can be predicted without the context, so the unfiltered benchmark is too easy a test of contextual hallucination detection. Compared with Table~\ref{tab:det-auc}, every proxy detector beats the text-only baselines on RAGTruth QA, where hallucinated answers are longer; on HaluBench and TRIVIA+, A$_7$ alone stays within 0.04 of TF--IDF, whereas A$_{64}$ and every set with hidden states exceed it by more than 0.1.

\begin{table}[t]
\centering
\footnotesize
\setlength{\tabcolsep}{3.5pt}
\begin{tabular}{lcccc}
\toprule
& RAGTruth & & \multicolumn{2}{c}{HaluBench}\\
\cmidrule(lr){4-5}
Baseline & QA & TRIVIA+ & filtered & unfiltered\\
\midrule
\multicolumn{5}{@{}l}{\emph{AUROC}}\\
Words & .702 & .574 & .580 & .727\\
Characters & .720 & .598 & .568 & .725\\
TF--IDF & .726 & .658 & .663 & .842\\
\midrule
\multicolumn{5}{@{}l}{\emph{F1}}\\
Words & .408 & .515 & .631 & .781\\
Characters & .426 & .515 & .628 & .785\\
TF--IDF & .390 & .529 & .642 & .755\\
\bottomrule
\end{tabular}
\caption{Text-only baselines on the test splits of Table~\ref{tab:det-data}: answer length in words or characters, and TF--IDF of the answer alone. The last column uses a split of HaluBench built the same way without the length filtering (2,231 test answers, 1,075 hallucinated). The 95\% group-bootstrap intervals of AUROC are at most $\pm0.05$.}
\label{tab:text-baselines}
\end{table}

\FloatBarrier
\section{Proxy detection: details}
\label{app:detection}
\subsection{Feature recipes}
\label{app:recipes}

We format each input using the proxy model's native chat template, retain the full context, question, and answer without truncation, and run inference with thinking mode disabled. We selected the feature subsets for A$_7$ and L$_9$ and the pooling and dimensionality reduction settings for H$_{64}$ once, independently for each block, using grouped five-fold cross-validation on the TRIVIA+ training split with Qwen3-4B. Because the benchmarks can share source questions and contexts, we first excluded 61 TRIVIA+ training groups containing examples whose parts exactly matched the corresponding field of a test example in another benchmark. 
This left 2,079 training examples for the search and avoided using overlapping test content to select subsequently evaluated configurations. For each block, we chose the variant with the highest mean AUROC across the five folds, breaking ties in favor of fewer features, simpler representations, and smaller $C$. 
These feature definitions and preprocessing choices are then reused across all other datasets and proxy models without repeating the selection.

\paragraph{Cheap attention features (A$_7$).}
The heads of each proxy are ranked once by their induction score, without hallucination labels, and A$_7$ reads the first $K$ heads of this ranking. 

The induction score of a head is its mean attention weight from each token of a repeated sequence to the token that followed its first occurrence, the weight an induction head puts on the token it copies \citep{olsson2022induction}, averaged over 128 sequences (seed 42), each a beginning-of-sequence token followed by 64 random tokens twice. The ranking is a one-time preparation step and is not part of the measured extraction time. 

The 22 candidate features of Section~\ref{sec:in-domain}, each averaged over the selected heads, were searched with $K\in\{1,2,4,8,16,32\}$ by recursive  elimination of one feature at a time. The seven kept features defined in Appendix~\ref{app:stats} are the maximum weight (mean, standard deviation, and maximum over answer tokens), effective support (minimum), entropy normalized by $\log n_i$ (mean and standard deviation), and the BOS sink rate. 
BOS is typically the top-1 sink of the selected heads.

Validation selects $K=32$, 16, 32, and 16 on RAGTruth QA for the 0.6B, 1.7B, 4B, and 8B proxies correspondingly, and $K=1$ on HaluBench with every proxy and on TRIVIA+ except with 8B ($K=16$).

\paragraph{All-head sink values (A$_{64}$).}
For each head, we divide each column sum by the number of answer rows that see its key, $\bar c_j=c_j/\min(R,L-j)$, and keep its $k$ largest values in descending order, without their positions. Concatenated over all heads and layers, standardized, and projected onto 64 principal components, they form A$_{64}$. The number $k\in\{1,\dots,10\}$ is chosen jointly with a provisional regularization of A$_{64}$ alone by five-fold grouped cross-validation on the training split, with the scaler and projection refit in every fold; the chosen $k$ is 1--4, except 9 for TRIVIA+ with 1.7B. A$_{64}$H$_{64}$ keeps that $k$ and reuses the projection of H$_{64}$; the final regularization and threshold of A$_{64}$ and A$_{64}$H$_{64}$ are chosen on validation as for every other set.

\paragraph{Compact hidden states (H$_{64}$).}
The block takes decoder layer 14 for 0.6B and 1.7B and layer 18 for 4B and 8B (counting from 1; hidden sizes 1024, 2048, 2560, and 4096), concatenates the per-coordinate mean and population standard deviation over answer tokens, standardizes them, projects them onto the first 64 components of a randomized PCA, and standardizes again; the scalers and the PCA are fit on the training split of each dataset and proxy. The selection compared three poolings (mean; mean and standard deviation; mean, standard deviation, minimum, and maximum) with 16, 32, 64, 128, 256, or 512 principal components or no projection. The layer itself is not selected: following prior work on hidden-state hallucination signals \citep{azaria2023internal,chen2024inside}, we take a middle layer.

\paragraph{Logit features (L$_9$).}
The 224 candidate features, the four pooled features of each statistic of Table~\ref{tab:logits}, were searched by recursive feature elimination as for A$_7$. The block keeps the standard deviation and minimum of the Zipf slope, the minimum Jensen--Shannon divergence, the mean and minimum answer-token probability, the minimum of the fourth-largest probability, and the maxima of the 7th, 10th, and 49th largest, 9 in total.

\subsection{Logit features}
\label{app:logits}
The logit block reads the proxy's output logits in the same forward as the attention statistics. A hook on the final normalized hidden states applies the language-model head only at the positions that predict answer tokens, plus the one before them, so the full vocabulary distribution is never formed for the prompt. Let $q_t$ be the next-token distribution at the position that predicts answer token $y_t$, $V$ the vocabulary, and $q_t^{(r)}$ the $r$-th largest probability of $q_t$. Table~\ref{tab:logits} defines the seven per-token statistics. Their mean, standard deviation, minimum, and maximum over the answer tokens, taken for the top-$k$ probabilities separately at each rank $r\le50$, are the 224 candidate features of Appendix~\ref{app:recipes}.

\begin{table*}[t]
\centering
\footnotesize
\setlength{\tabcolsep}{4pt}
\renewcommand{\arraystretch}{1.2}
\begin{tabular}{@{}>{\raggedright\arraybackslash}p{0.22\textwidth}>{\raggedright\arraybackslash}p{0.62\textwidth}c@{}}
\toprule
Statistic & Definition (per answer token $t$) & Shape\\
\midrule
Answer-token probability & $q_t(y_t)$ & $R$\\
Answer-token rank & $1+|\{v\in V: q_t(v)>q_t(y_t)\}|$ & $R$\\
Top-$k$ probabilities & $q_t^{(r)}$ for each rank $r=1,\dots,50$ & $R\times50$\\
Entropy & $-\sum_{v\in V}q_t(v)\log q_t(v)$ & $R$\\
Effective vocabulary size & $\bigl(\sum_{v\in V}q_t(v)^2\bigr)^{-1}$ & $R$\\
Zipf slope & least-squares slope of $\log q_t^{(r)}$ against $\log r$, $r=1,\dots,50$ & $R$\\
Jensen--Shannon divergence & $\mathrm{JS}(q_t\,\|\,q_{t-1})$ between consecutive positions; $q_{t-1}$ of the first answer token is the last prompt position & $R$\\
\bottomrule
\end{tabular}
\caption{Logit statistics computed per answer token from the proxy's next-token distribution. Shape: output for one sequence with $R$ answer tokens.}
\label{tab:logits}
\end{table*}

\subsection{Training and significance tests}
\label{app:det-tests}
The regression uses \texttt{liblinear} with balanced class weights. For each dataset, proxy, and feature set, $C\in\{0.001,0.01,0.1,1,10\}$ is chosen by validation AUROC and the threshold by validation F1, with ties broken toward smaller $K$ and $C$ and, for the threshold, toward the higher one (Section~\ref{sec:selection}). Models are not refit on training and validation data together. To test an addition of an attention block, we draw 100{,}000 paired bootstrap resamples of test groups (Table~\ref{tab:det-data}; seed 42), keeping models, regularization, and thresholds fixed. The two-sided $p$-value of an observed difference $\Delta$ is the share of centered bootstrap differences at least as large in absolute value, $p=(1+\#\{b:|\Delta^*_b-\Delta|\ge|\Delta|\})/(B+1)$ with $B=100{,}000$, and we Holm-correct these $p$-values over the 12 dataset--proxy pairs of each addition. With only 150--546 test groups, the intervals are wide, so only large gains stay significant. Table~\ref{tab:det-sig} tests each addition of A$_7$, and Table~\ref{tab:sink-sig} tests H$_{64}{\to}$A$_{64}$H$_{64}$.

\subsection{Full results}
\label{app:det-results}
Table~\ref{tab:det-auc} gives the test AUROC of every detector of Table~\ref{tab:det-f1}. Tables~\ref{tab:det-sig} and~\ref{tab:sink-sig} test the additions of A$_7$ and A$_{64}$ (Appendix~\ref{app:det-tests}); bold marks the gains that stay significant after correction over the 12 pairs of an addition. The significant gains fall where each block helps most, and no tested addition significantly lowers F1.

\begin{table*}[t]
\centering
\small
\setlength{\tabcolsep}{3.6pt}
\begin{tabular}{lr*{12}{c}}
\toprule
& & \multicolumn{4}{c}{RAGTruth QA} & \multicolumn{4}{c}{HaluBench} & \multicolumn{4}{c}{TRIVIA+}\\
\cmidrule(lr){3-6}\cmidrule(lr){7-10}\cmidrule(lr){11-14}
Features & $N$ & 0.6B & 1.7B & 4B & 8B & 0.6B & 1.7B & 4B & 8B & 0.6B & 1.7B & 4B & 8B\\
\midrule
A$_7$ & 7 & .852 & .847 & .838 & .856 & .627 & .636 & .652 & .660 & .642 & .653 & .666 & .645\\
\addlinespace[1pt]
L$_9$ & 9 & .764 & .765 & .793 & .765 & .654 & .646 & .661 & .680 & .684 & .661 & .696 & .703\\
A$_7$L$_9$ & 16 & .852 & .849 & .859 & .859 & .663 & .667 & .680 & .693 & .686 & .682 & .706 & .707\\
\addlinespace[1pt]
H$_{64}$ & 64 & .864 & .855 & .869 & .862 & .792 & \underline{.824} & .846 & .846 & .775 & .786 & \underline{.807} & .804\\
A$_7$H$_{64}$ & 71 & \underline{.871} & \underline{.861} & .866 & .872 & .795 & \underline{.824} & .846 & .847 & .777 & .789 & .806 & .793\\
\addlinespace[1pt]
H$_{64}$L$_9$ & 73 & .865 & .856 & \underline{.876} & .867 & .797 & .823 & .848 & .851 & .776 & .791 & .804 & .804\\
A$_7$H$_{64}$L$_9$ & 80 & \textbf{.873} & \underline{.861} & \textbf{.877} & \underline{.875} & .800 & .822 & .848 & .851 & \underline{.781} & \textbf{.796} & .799 & \underline{.813}\\
\midrule
A$_{64}$ & 64 & .866 & \textbf{.873} & .872 & \textbf{.880} & \underline{.801} & .821 & \underline{.871} & \underline{.877} & .780 & .774 & .801 & .803\\
A$_{64}$H$_{64}$ & 128 & .866 & .857 & .875 & .874 & \textbf{.814} & \textbf{.846} & \textbf{.877} & \textbf{.881} & \textbf{.796} & \underline{.792} & \textbf{.820} & \textbf{.824}\\
\midrule
\multicolumn{2}{@{}l}{Best LLM judge} & \multicolumn{4}{c}{.909} & \multicolumn{4}{c}{.859} & \multicolumn{4}{c}{.786}\\
\bottomrule
\end{tabular}
\caption{Test AUROC of the detectors of Table~\ref{tab:det-f1}. \textbf{Bold}: best in column; \underline{underlined}: second best (ties marked alike). Bottom: the best of nine LLM judges of 4--9B parameters, Qwen3.5-9B on every dataset (Table~\ref{tab:judges}).}
\label{tab:det-auc}
\end{table*}

\begin{table*}[t]
\centering
\small
\setlength{\tabcolsep}{2.8pt}
\begin{tabular}{ll*{12}{c}}
\toprule
& & \multicolumn{4}{c}{RAGTruth QA} & \multicolumn{4}{c}{HaluBench} & \multicolumn{4}{c}{TRIVIA+}\\
\cmidrule(lr){3-6}\cmidrule(lr){7-10}\cmidrule(lr){11-14}
& Addition & 0.6B & 1.7B & 4B & 8B & 0.6B & 1.7B & 4B & 8B & 0.6B & 1.7B & 4B & 8B\\
\midrule
\multirow{3}{*}{$\Delta$F1} & H$_{64}{\to}$A$_7$H$_{64}$ & +.080 & +.022 & $-$.020 & +.043 & +.001 & $-$.005 & +.002 & +.004 & $-$.014 & .000 & +.005 & $-$.025\\
& L$_9{\to}$A$_7$L$_9$ & +.126 & +.088 & +.074 & +.114 & +.009 & +.015 & +.014 & +.009 & +.011 & +.038 & +.025 & +.002\\
& H$_{64}$L$_9{\to}$A$_7$H$_{64}$L$_9$ & +.007 & +.005 & $-$.016 & +.030 & +.006 & $-$.010 & +.006 & +.002 & +.042 & +.002 & +.005 & +.020\\
\midrule
\multirow{3}{*}{$p$, Holm-12} & H$_{64}{\to}$A$_7$H$_{64}$ & \textbf{.005} & .557 & .605 & \textbf{.017} & 1 & 1 & 1 & 1 & 1 & 1 & 1 & 1\\
& L$_9{\to}$A$_7$L$_9$ & $\boldsymbol{<}$\textbf{.001} & \textbf{.004} & \textbf{.002} & $\boldsymbol{<}$\textbf{.001} & 1 & 1 & .503 & 1 & 1 & .317 & 1 & 1\\
& H$_{64}$L$_9{\to}$A$_7$H$_{64}$L$_9$ & 1 & 1 & .885 & .065 & 1 & .458 & 1 & 1 & .174 & 1 & 1 & .477\\
\bottomrule
\end{tabular}
\caption{Change in test F1 when A$_7$ is added, with two-sided $p$-values from 100{,}000 paired bootstrap resamples of test groups (seed 42; models, regularization, and thresholds fixed), Holm-corrected over the 12 dataset--proxy pairs of an addition (Holm-12). A gain is significant if $\Delta>0$ and the corrected $p<0.05$; \textbf{bold}: significant gain. No addition gives a significant loss.}
\label{tab:det-sig}
\end{table*}

\begin{table*}[t]
\centering
\small
\setlength{\tabcolsep}{3.4pt}
\begin{tabular}{ll*{12}{c}}
\toprule
& & \multicolumn{4}{c}{RAGTruth QA} & \multicolumn{4}{c}{HaluBench} & \multicolumn{4}{c}{TRIVIA+}\\
\cmidrule(lr){3-6}\cmidrule(lr){7-10}\cmidrule(lr){11-14}
& Addition & 0.6B & 1.7B & 4B & 8B & 0.6B & 1.7B & 4B & 8B & 0.6B & 1.7B & 4B & 8B\\
\midrule
$\Delta$F1 & H$_{64}{\to}$A$_{64}$H$_{64}$ & +.044 & $-$.030 & $-$.001 & +.037 & +.027 & +.012 & +.035 & +.050 & +.009 & +.002 & +.040 & +.001\\
$p$, Holm-12 & H$_{64}{\to}$A$_{64}$H$_{64}$ & .201 & .324 & 1 & .269 & .198 & 1 & \textbf{.027} & $\boldsymbol{<}$\textbf{.001} & 1 & 1 & .189 & 1\\
\bottomrule
\end{tabular}
\caption{Change in test F1 when A$_{64}$ is added to H$_{64}$ (Section~\ref{sec:sink64}), with two-sided $p$-values from 100{,}000 paired bootstrap resamples of test groups as in Table~\ref{tab:det-sig}, Holm-corrected over the 12 dataset--proxy pairs (Holm-12). \textbf{Bold}: significant gain. No change is a significant loss.}
\label{tab:sink-sig}
\end{table*}

\subsection{LLM judges}
\label{app:judges}
We follow a typical protocol for LLM judges.
Each judge receives the context, question, and candidate answer
using one of the four prompt templates in
Appendix~\ref{app:prompts}. With thinking disabled, it produces
one token, restricted to \texttt{True} (hallucinated) or
\texttt{False} (supported). We use the probability assigned to
\texttt{True} as the hallucination score. The templates contain
either no demonstrations or eight labeled demonstrations.
For each of the nine judges, we select the template using the
three training splits combined and select the decision threshold
by validation F1, as for the proxy detectors.

The four Qwen3 proxy models are prompted as judges with each template, and scored like the judges.
For each model we keep the template with the best mean validation AUROC (Strict few-shot for 4B and 8B, Claims for 0.6B and 1.7B). 
Bespoke-MiniCheck-7B \citep{tang2024minicheck} and
HHEM-2.1-Open \citep{li2024hhem} estimate the probability
that a premise supports a claim. We use one minus this
probability as the hallucination score.
MiniCheck splits long premises into chunks and keeps the highest support, and we do the same for HHEM above 8,192 tokens. MiniCheck is trained on ANLI and synthetic data; the training data of HHEM is not disclosed.

The nine judges are Qwen3.5-9B \citep{qwen2026qwen35}, Gemma-4-E4B \citep{gemma2026gemma4}, Granite-4.2-8B \citep{ibm2026granite42}, Ministral-3-8B \citep{liu2026ministral3}, Falcon-H1-7B \citep{zuo2025falconh1}, Llama-3.1-8B \citep{dubey2024llama3}, GLM-4-9B \citep{glm2024chatglm}, Phi-4-mini \citep{abouelenin2025phi4mini}, and Command-R7B \citep{cohere2024commandr7b}.
We report results for them in Table~\ref{tab:judges}. 

\begin{table*}[t]
\centering
\small
\setlength{\tabcolsep}{4pt}
\begin{tabular}{llrrrrrr}
\toprule
& & \multicolumn{3}{c}{AUROC} & \multicolumn{3}{c}{F1}\\
\cmidrule(lr){3-5}\cmidrule(lr){6-8}
Judge & Prompt & RAGTruth & HaluB. & TRIVIA+ & RAGTruth & HaluB. & TRIVIA+\\
\midrule
Qwen3.5-9B & Strict few-shot & .909 & .859 & .786 & .623 & .755 & .655\\
Gemma-4-E4B & Evidence & .878 & .855 & .774 & .617 & .751 & .621\\
Granite-4.2-8B & Strict & .843 & .848 & .755 & .547 & .753 & .645\\
Ministral-3-8B & Strict & .823 & .817 & .735 & .517 & .721 & .613\\
Falcon-H1-7B & Strict few-shot & .752 & .834 & .747 & .434 & .726 & .624\\
Llama-3.1-8B & Strict & .688 & .754 & .688 & .363 & .684 & .564\\
GLM-4-9B & Strict few-shot & .725 & .753 & .683 & .399 & .678 & .587\\
Phi-4-mini & Claims & .550 & .771 & .664 & .306 & .693 & .566\\
Command-R7B & Evidence & .501 & .604 & .624 & .302 & .634 & .536\\
\midrule
Qwen3-8B, zero/few-shot & Strict few-shot & .814 & .835 & .752 & .521 & .742 & .638\\
Qwen3-4B, zero/few-shot & Strict few-shot & .769 & .799 & .753 & .439 & .724 & .631\\
Qwen3-1.7B, zero/few-shot & Claims & .443 & .510 & .433 & .309 & .622 & .492\\
Qwen3-0.6B, zero/few-shot & Claims & .469 & .473 & .495 & .306 & .622 & .516\\
Bespoke-MiniCheck-7B & -- & .881 & .822 & .751 & .578 & .717 & .616\\
HHEM-2.1-Open & -- & .852 & .703 & .726 & .563 & .663 & .604\\
\midrule
Qwen3-0.6B, A$_7$H$_{64}$L$_9$ & -- & .873 & .800 & .781 & .621 & .702 & .641\\
Qwen3-4B, A$_7$H$_{64}$L$_9$ & -- & .877 & .848 & .799 & .528 & .755 & .646\\
Qwen3-0.6B, A$_{64}$H$_{64}$ & -- & .866 & .814 & .796 & .601 & .724 & .636\\
Qwen3-4B, A$_{64}$H$_{64}$ & -- & .875 & .877 & .820 & .566 & .776 & .674\\
\bottomrule
\end{tabular}
\caption{LLM judges on the test splits of Table~\ref{tab:det-data} with F1 at the threshold selected on validation. 
Top: nine instruction-tuned LLMs prompted as judges.
Middle: the four Qwen3 proxy backbones prompted as judges and two specialized grounding checkers.
Bottom: our proxy detectors for reference.}
\label{tab:judges}
\end{table*}

\FloatBarrier
\section{Cross-domain transfer on HaluBench}
\label{app:extra}
\label{app:transfer}
Section~\ref{sec:evaluation} fits and tests each detector within one dataset. Attention detectors that use few or no labels, however, are meant to work on domains they were not tuned on. We test whether they do across the six sources of HaluBench, with five published detectors: Lookback Lens, LLM-Check, RAUQ, and INTRYGUE, whose statistics our kernels compute (Table~\ref{tab:stats}), and TOHA, which needs the materialized response block, plus the sink-rate term of INTRYGUE alone.

\paragraph{Protocol.}
Qwen3-8B reads the context, question, and existing answer of every example in our filtered HaluBench split (Appendix~\ref{app:data}), teacher-forced with its chat template and thinking off. One forward extracts the features of all methods, with our kernels for the statistics of Table~\ref{tab:stats} and the materialized response block for TOHA (T4); we then fit and test them on CPU. Each of the six sources has 138--166 test examples. RAUQ needs no labels \citep{vazhentsev2025rauq}. INTRYGUE \citep{kuleshova2026intrygue}, its sink-rate term alone, TOHA \citep{bazarova2025toha}, LLM-Check \citep{sriramanan2024llmcheck}, and our variant of Lookback Lens, which averages the lookback ratios of at most 10 heads instead of training a probe on all heads \citep{chuang2024lookback}, choose one setting by AUROC on the validation split of the fitting sources, about 150 labels per source: the number of induction heads, the layer, or the heads.

Each method is fit on one source and tested on all six, fit on the five sources other than the test domain (leave-one-source-out), and fit on all six. Table~\ref{tab:transfer} gives the test AUROC of every setting; 95\% bootstrap intervals use 1,000 resamples of individual test examples rather than of the context groups used in Section~\ref{sec:evaluation}, so they are somewhat narrower than group intervals would be.

\begin{table*}[t]
\centering
\small
\setlength{\tabcolsep}{5pt}
\begin{tabular}{@{}ll*{6}{c}c@{}}
\toprule
& & \multicolumn{6}{c}{Test domain} & \\
\cmidrule(lr){3-8}
Fit on & Method & RAGTruth & CovidQA & PubMedQA & FinanceBench & DROP & HaluEval & Mean\\
\midrule
\multicolumn{2}{@{}l}{RAUQ (no fitting)} & .58 & .67 & .66 & .52 & .66 & .85 & .66\\
\midrule
\multirow{5}{*}{RAGTruth} & INTRYGUE & \cellcolor{black!12}\textbf{.89} & \textbf{.77} & .55 & .48 & .41 & \textbf{.92} & .62\\
 & Sink rate & \cellcolor{black!12}.80 & .72 & .53 & .49 & .56 & .91 & \textbf{.64}\\
 & LLM-Check & \cellcolor{black!12}.59 & .58 & \textbf{.62} & \textbf{.53} & .53 & .75 & .60\\
 & TOHA & \cellcolor{black!12}.68 & .59 & .44 & .46 & \textbf{.61} & .80 & .58\\
 & Lookback Lens & \cellcolor{black!12}.88 & .56 & .47 & .51 & .49 & .73 & .55\\
\midrule
\multirow{5}{*}{CovidQA} & INTRYGUE & \textbf{.89} & \cellcolor{black!12}\textbf{.77} & .55 & \textbf{.48} & .41 & \textbf{.92} & \textbf{.65}\\
 & Sink rate & .51 & \cellcolor{black!12}.62 & .52 & .44 & .55 & .78 & .56\\
 & LLM-Check & .44 & \cellcolor{black!12}.55 & \textbf{.61} & .45 & .56 & .59 & .53\\
 & TOHA & .68 & \cellcolor{black!12}.59 & .44 & .46 & \textbf{.61} & .80 & .60\\
 & Lookback Lens & .31 & \cellcolor{black!12}.65 & .40 & .47 & .53 & .57 & .45\\
\midrule
\multirow{5}{*}{PubMedQA} & INTRYGUE & \textbf{.89} & \textbf{.75} & \cellcolor{black!12}.55 & .47 & .40 & \textbf{.92} & \textbf{.69}\\
 & Sink rate & .51 & .62 & \cellcolor{black!12}.52 & .44 & .54 & .82 & .59\\
 & LLM-Check & .65 & .60 & \cellcolor{black!12}\textbf{.68} & .44 & \textbf{.59} & .71 & .60\\
 & TOHA & .39 & .59 & \cellcolor{black!12}.50 & \textbf{.53} & .51 & .43 & .49\\
 & Lookback Lens & .80 & .40 & \cellcolor{black!12}.53 & .52 & .50 & .48 & .54\\
\midrule
\multirow{5}{*}{FinanceBench} & INTRYGUE & \textbf{.82} & \textbf{.70} & \textbf{.54} & \cellcolor{black!12}.47 & .38 & \textbf{.86} & \textbf{.66}\\
 & Sink rate & .51 & .62 & .52 & \cellcolor{black!12}.44 & .54 & .82 & .60\\
 & LLM-Check & .42 & .61 & .49 & \cellcolor{black!12}.45 & .56 & .78 & .57\\
 & TOHA & .68 & .54 & .44 & \cellcolor{black!12}.48 & \textbf{.57} & .55 & .55\\
 & Lookback Lens & .68 & .45 & .41 & \cellcolor{black!12}\textbf{.51} & .54 & .37 & .49\\
\midrule
\multirow{5}{*}{DROP} & INTRYGUE & \textbf{.82} & \textbf{.70} & \textbf{.54} & .47 & \cellcolor{black!12}.39 & \textbf{.87} & \textbf{.68}\\
 & Sink rate & .72 & .68 & \textbf{.54} & .44 & \cellcolor{black!12}.56 & .86 & .65\\
 & LLM-Check & .44 & .63 & .45 & .47 & \cellcolor{black!12}.58 & .81 & .56\\
 & TOHA & .66 & .52 & .41 & .49 & \cellcolor{black!12}\textbf{.61} & .59 & .54\\
 & Lookback Lens & .66 & .40 & .37 & \textbf{.53} & \cellcolor{black!12}.58 & .41 & .47\\
\midrule
\multirow{5}{*}{HaluEval} & INTRYGUE & \textbf{.89} & \textbf{.75} & \textbf{.55} & .47 & .40 & \cellcolor{black!12}.92 & \textbf{.61}\\
 & Sink rate & .73 & .74 & .51 & .49 & .56 & \cellcolor{black!12}\textbf{.93} & \textbf{.61}\\
 & LLM-Check & .59 & .63 & .50 & .46 & .57 & \cellcolor{black!12}.88 & .55\\
 & TOHA & .68 & .59 & .44 & .46 & \textbf{.60} & \cellcolor{black!12}.81 & .56\\
 & Lookback Lens & .79 & .52 & .39 & \textbf{.52} & .42 & \cellcolor{black!12}.82 & .53\\
\midrule
\multirow{5}{*}{Other five} & INTRYGUE & \textbf{.89} & \textbf{.77} & \textbf{.55} & .48 & .41 & \textbf{.92} & \textbf{.67}\\
 & Sink rate & .80 & .72 & .53 & \textbf{.49} & .56 & .91 & \textbf{.67}\\
 & LLM-Check & .59 & .63 & .45 & .46 & .56 & .71 & .57\\
 & TOHA & .68 & .59 & .44 & .46 & \textbf{.60} & .55 & .55\\
 & Lookback Lens & .71 & .54 & .40 & .47 & .53 & .52 & .53\\
\midrule
\multirow{5}{*}{All six} & INTRYGUE & \textbf{.89} & \textbf{.77} & \textbf{.55} & .48 & .41 & \textbf{.92} & \textbf{.67}\\
 & Sink rate & .80 & .72 & .53 & .49 & .56 & .91 & \textbf{.67}\\
 & LLM-Check & .59 & .63 & .50 & .46 & .57 & .88 & .60\\
 & TOHA & .68 & .59 & .44 & .46 & \textbf{.60} & .81 & .60\\
 & Lookback Lens & .36 & .54 & .40 & \textbf{.52} & \textbf{.60} & .48 & .48\\
\bottomrule
\end{tabular}
\caption{Cross-domain transfer on HaluBench (Qwen3-8B reader): test AUROC of each method fit on the source of its row group and tested on each source. Shaded: same source. \emph{Other five}: fit on the five sources other than the test domain (leave-one-source-out); \emph{All six}: fit on all six sources (pooled). Mean: over the five other test domains for a single fitting source, over all six for the last two groups. RAUQ fits nothing and scores the same for every fitting source. \textbf{Bold}: best method of the group in each column (ties marked alike). Each test domain has 138--166 examples; the 95\% bootstrap intervals span $\pm0.09$ at the median ($\pm0.03$ to $\pm0.12$).}
\label{tab:transfer}
\end{table*}

\paragraph{Results.}
RAUQ and INTRYGUE transfer; detectors that select heads or a layer on one source do not. Averaged over the six fitting sources, RAUQ and INTRYGUE lose at most 0.01 AUROC from the fitting source to the other five, and the sink rate 0.04, whereas LLM-Check loses 0.05, TOHA 0.06, and Lookback Lens 0.15, which falls to chance (0.51). Fitting on more sources does not help much, since each detector still selects only one setting. Two domains defeat every method: a fixed-direction score cannot adapt where its direction reverses, so INTRYGUE stays at 0.38--0.41 on DROP from every source, and no method exceeds 0.53 on FinanceBench.

\FloatBarrier
\section{LLM judge prompts}
\label{app:prompts}
Each judge of Table~\ref{tab:judges} reads one of the four templates below, given verbatim. \texttt{\#\#\# system} and \texttt{\#\#\# user} open the system and user messages of the judge's chat template; \texttt{\{context\}}, \texttt{\{question\}}, and \texttt{\{response\}} are filled with the example. Every template asks for the verdict \texttt{True} (hallucinated) or \texttt{False} (supported).

\paragraph{Evidence} (Gemma-4-E4B, Command-R7B).
\begin{Verbatim}[fontsize=\scriptsize,breaklines,frame=single,framesep=2pt]
### user
You are assessing a candidate answer to a question. You are given the question, reference evidence, and one candidate answer.

# Question
{question}

# Reference evidence
{context}

# Candidate answer
{response}

Definition. A hallucinated answer contains information that is incorrect, fabricated, or not supported by the evidence: for example, it misreads the question, reverses a finding, invents a mechanism, or adds specifics absent from the source. A supported answer is fully consistent with the evidence.

Is the candidate answer hallucinated? Answer with exactly one word: True if it is hallucinated or unsupported, False if it is fully supported.
\end{Verbatim}

\paragraph{Claims} (Phi-4-mini).
\begin{Verbatim}[fontsize=\scriptsize,breaklines,frame=single,framesep=2pt]
### system
You check model responses against a source document. Use only the document, never outside knowledge. A response passes only if every factual claim in it is stated in, or directly entailed by, the document. Any invented, altered, or unverifiable detail makes it fail, even if the rest is correct. Leaving information out does not make it fail.
### user
Document:
"""
{context}
"""

Request: {question}

Response:
"""
{response}
"""

Does the response contain any claim that the document does not support? Answer True (it does, the response is hallucinated) or False (every claim is supported).
\end{Verbatim}

\paragraph{Strict} (Granite-4.2-8B, Ministral-3-8B, Llama-3.1-8B) is Strict few-shot without its block of examples, from \texttt{Examples:} up to \texttt{Output format:}.

\paragraph{Strict few-shot} (Qwen3.5-9B, Falcon-H1-7B, GLM-4-9B). The prompt contains eight author-written illustrative examples: four hallucinated and four faithful.

\begin{Verbatim}[fontsize=\scriptsize,breaklines,frame=single,framesep=2pt]
### system
You are an expert evaluator specialized in detecting hallucinations. Use only the CONTEXT, never outside knowledge.

Your task is to determine whether the RESPONSE contains any information that the CONTEXT does not support, or answers the QUESTION incorrectly according to the CONTEXT.

A response is faithful only if every factual claim in it is stated in, or directly entailed by, the context. Any invented, altered, or unverifiable detail makes it hallucinated, even if the rest of the response is correct.

True means HALLUCINATED:
- At least one claim is not stated in or entailed by the context, including plausible but unstated details and claims more specific than the context allows.
- The response invents or alters entities, numbers, dates, policies, events, or relationships.
- The response gives a number, name, or date that differs from what the context implies for this question, including wrong counting or arithmetic over values in the context.
- The response answers a different question than the one asked (wrong entity, team, period, or unit).
- The response refuses to answer although the context clearly contains the answer.

False means FAITHFUL:
- Every factual claim is supported by the context and the response correctly answers the question.
- The response declines to answer because the context does not contain the required information; such a refusal is faithful.

Important rules:
- One unsupported or incorrect claim is enough to mark the response as hallucinated.
- Leaving information out is not a hallucination.
- A plausible or well-known fact is still hallucinated if the context does not support it.
- For short answers, recompute or re-locate the answer in the context and compare exactly.

Examples:

<context>The museum opened in 1998 and has three floors of galleries.</context>
<question>When did the museum open?</question>
<response>The museum opened in 1998 and attracts over a million visitors a year.</response>
Answer: True
(The visitor number is not in the context: an added unsupported detail.)

<context>In the first half the Hawks scored 14 points and the Owls scored 10. In the second half the Hawks scored 7 and the Owls scored 21.</context>
<question>How many points did the Owls score in total?</question>
<response>24</response>
Answer: True
(10 + 21 = 31: wrong arithmetic over the context.)

<context>Maria Lopez founded the company in 2004. Her brother Diego joined as CFO in 2010.</context>
<question>Who founded the company?</question>
<response>Diego Lopez founded the company.</response>
Answer: True
(The response names the wrong person, contradicting the context.)

<context>The trial found that the drug lowered blood pressure in 62% of patients but had no effect on heart rate.</context>
<question>What effect did the drug have on heart rate?</question>
<response>The drug noticeably slowed patients' heart rate.</response>
Answer: True
(The context says there was no effect: the response reverses the finding.)

<context>The bridge, completed in 1932, spans 503 meters and carries both trains and cars.</context>
<question>How long is the bridge?</question>
<response>It is about half a kilometer long, 503 meters.</response>
Answer: False
(A faithful paraphrase with an equivalent unit.)

<context>The library offers free Wi-Fi, study rooms, a children's section, and a cafe.</context>
<question>What services does the library offer?</question>
<response>The library has free Wi-Fi and study rooms.</response>
Answer: False
(Incomplete but every stated claim is supported; omissions are not hallucinations.)

<context>The report covers the company's sales in Europe for 2021.</context>
<question>What were the company's sales in Asia in 2021?</question>
<response>The context does not contain information about sales in Asia.</response>
Answer: False
(A correct refusal: the answer is not in the context.)

<context>Team A won 3 games in March and 5 games in April.</context>
<question>How many games did Team A win in March and April combined?</question>
<response>8</response>
Answer: False
(3 + 5 = 8: correctly derived from the context.)

Output format:
- Reply with exactly one word: True or False.
- True = the response is hallucinated; False = the response is faithful.
- Do not write any explanation, punctuation, quotes, tags, or other text before or after the word.
### user
<context>
{context}
</context>
<question>
{question}
</question>
<response>
{response}
</response>
Answer:
\end{Verbatim}

\section{Use of scientific artifacts}
\label{app:artifacts}
Table~\ref{tab:artifacts} lists the licences of the datasets and models we use. We use all of them for non-commercial research, consistent with their terms, and redistribute none of them; this matters for the non-commercial licences and for TRIVIA+, whose licence forbids derivatives. Our anonymized code is released under the MIT licence.

\begin{table}[h]
\centering
\footnotesize
\setlength{\tabcolsep}{3pt}
\begin{tabular}{@{}lll@{}}
\toprule
Artifact & Use & Licence\\
\midrule
RAGTruth & detection & MIT\\
HaluBench & detection & CC BY-NC 2.0\\
TRIVIA+ & detection & CC BY-NC-ND 4.0\\
PHANTOM & kernels & Apache 2.0\\
\midrule
Qwen3 0.6B--8B & proxy, judge & Apache 2.0\\
Qwen3.5-9B & judge & Apache 2.0\\
Gemma-4-E4B & judge & Apache 2.0\\
Granite-4.2-8B & judge & Apache 2.0\\
Ministral-3-8B & judge & Apache 2.0\\
Phi-4-mini & judge & MIT\\
Llama-3.1-8B & judge & Llama 3.1 License\\
Falcon-H1-7B & judge & Falcon LLM\\
GLM-4-9B & judge & GLM-4\\
Command-R7B & judge & CC BY-NC 4.0\\
Bespoke-MiniCheck-7B & checker & CC BY-NC 4.0\\
HHEM-2.1-Open & checker & Apache 2.0\\
\bottomrule
\end{tabular}
\caption{Licences of the artifacts we use, as stated by their releases; Falcon LLM and GLM-4 are the custom licences of these models. Detection: Section~\ref{sec:evaluation}; kernels: Section~\ref{sec:kernel-eval}; judges and grounding checkers (checker): Appendix~\ref{app:judges}, where the models are cited.}
\label{tab:artifacts}
\end{table}

AI assistants were used to proofread and refine the text for clarity.

\end{document}